\documentclass{article}
\usepackage{ijcai22}

\usepackage{times}
\usepackage{soul}
\usepackage{url}
\usepackage[hidelinks]{hyperref}
\usepackage[utf8]{inputenc}
\usepackage[small]{caption}
\usepackage{graphicx}
\usepackage{amsmath}
\usepackage{amsthm}
\usepackage{booktabs}
\usepackage{algorithm}
\usepackage{algorithmicx}
\usepackage{makecell}

\usepackage{algpseudocode}

\usepackage{amssymb}

\title{Imitation Learning by State-Only Distribution Matching}

\author{
Damian Boborzi\footnote{Both Authors contributed equally}$^{1}$
\and
Christoph-Nikolas Straehle$^{*2}$\and
Jens S. Buchner$^{3}$\And
Lars Mikelsons$^1$
\affiliations
$^1$Augsburg University\\
$^2$Bosch Center for Artificial Intelligence\\
$^3$ETAS GmbH\\
\emails
damian.boborzi@uni-a.de
}

\begin{document}

\maketitle

\begin{abstract}
Imitation Learning from observation describes policy learning in a similar way to human learning. An agent's policy is trained by observing an expert performing a task. While many state-only imitation learning approaches are based on adversarial imitation learning, one main drawback is that adversarial training is often unstable and lacks a reliable convergence estimator. If the true environment reward is unknown and cannot be used to select the best-performing model, this can result in bad real-world policy performance. We propose a non-adversarial learning-from-observations approach, together with an interpretable convergence and performance metric. 

Our training objective minimizes the Kulback-Leibler divergence (KLD) between the policy and expert state transition trajectories which can be optimized in a non-adversarial fashion. Such methods demonstrate improved robustness when learned density models guide the optimization. We further improve the sample efficiency by rewriting the KLD minimization as the Soft Actor Critic objective based on a modified reward using additional density models that estimate the environment's forward and backward dynamics. Finally, we evaluate the effectiveness of our approach on well-known continuous control environments and show state-of-the-art performance while having a reliable performance estimator compared to several recent learning-from-observation methods.
\end{abstract}

\section{INTRODUCTION}

Imitation learning (IL) describes methods that learn optimal behavior that is represented by a collection of expert demonstrations. While in standard reinforcement learning (RL), the agent is trained on environment feedback using a reward signal, IL can alleviate the problem of designing effective reward functions. This is particularly useful for tasks where demonstrations are more accessible than designing a reward function. One popular example is to train traffic agents in a simulation to mimic real-world road users \cite{kuefler2017imitating}. 

Learning-from-demonstrations (LfD) describes IL approaches that require state-action pairs from expert demonstrations \cite{GAIL}. While actions can guide policy learning, it might be very costly or even impossible to collect actions alongside state demonstrations in many real-world setups. For example, when expert demonstrations are available as video recordings without additional sensor signals. One example of such a setup is training traffic agents in a simulation, where the expert data contains recordings of traffic in bird's eye view \cite{kuefler2017imitating}. No direct information on the vehicle physics, throttle, and steering angle is available. Another example is teaching a robot to pick, move, and place objects based on human demonstrations \cite{osa2018an}. In such scenarios, actions have to be estimated based on sometimes incomplete information to train an agent to imitate the observed behavior.  

Alternatively, learning-from-observations (LfO) performs state-only IL and trains an agent without actions being available in the expert dataset \cite{torabi2019recent}. While LfO is a more challenging task than LfD, it can be more practical in case of incomplete data sources. Learning an additional environment model may help to infer actions based on expert observations and the learned environment dynamics \cite{torabi2018behavioral}. Like LfD, distribution matching based on an adversarial setup is commonly used in LfO \cite{torabi2018generative}. In adversarial imitation learning (AIL), a policy is trained using an adversarial discriminator, which is used to estimate a reward that guides policy training. While AIL methods obtain better performing agents than supervised methods like behavioral cloning (BC) using less data, adversarial training often has stability issues \cite{miyato2018spectral} and under some conditions is not guaranteed to converge \cite{Jin2020What}. 
Additionally, estimating the performance of a trained policy without access to the environment reward can be very challenging. While the duality gap \cite{grnarova2019a,Sidheekh2021OnDG} is a convergence metric suited for GAN based methods, it is difficult to use in the AIL setup since it relies on the gradient of the generator for an optimization process. In the AIL setup, the generator consists of the policy and the environment and therefore the gradient is difficult to estimate with black box environments. As an alternative for AIL setups the predicted reward (discriminator output) or the policy loss can be used to estimate the performance.  

To address the limitations of AIL, we propose a state-only distribution matching method that learns a policy in a non-adversarial way. We optimize the Kulback-Leibler divergence (KLD) between the actionless policy and expert trajectories by minimizing the KLD of the conditional state transition distribution of the policy and the expert for all time steps. We estimate the expert state transition distribution using normalizing flows, which can be trained offline using the expert dataset. Thus, stability issues arising from the min-max adversarial optimization in AIL methods can be avoided. This objective is similar to FORM \cite{jaegle2021imitation}, which was shown to be more stable in the presence of task-irrelevant features.

While maximum entropy RL methods \cite{Ziebart2010} can improve policy training by increasing the exploration of the agent, they also add a bias if being used to minimize the proposed KLD. To match the transition distributions of the policy and the expert exactly, the state-next-state distribution of the policy is expanded into policy entropy, forward dynamics and inverse action model of the environment. It has been shown that such dynamic models can improve the convergence \cite{zhu2020off} and are well suited to infer actions not available in the dataset \cite{torabi2018behavioral}. We model these distributions using normalizing flow models which have been demonstrated to perform very well on learning complex probability distributions \cite{papamakarios2019normalizing}. Combining all estimates results in an interpretable reward that can be used together with standard maximum entropy RL methods \cite{haarnoja2018SAC}. The optimization based on the KLD provides a reliable convergence metric of the training and a good estimator for policy performance. 

As contributions we derive SOIL-TDM (State Only Imitation Learning by Trajectory Distribution Matching), a non-adversarial LfO method which minimizes the KLD between the conditional state transition distributions of the policy and the expert using maximum entropy RL. We show the convergence of the proposed method using off-policy samples from a replay buffer. We develop a practical algorithm based on the SOIL-TDM objective and demonstrate its effectiveness to measure its convergence compared to several other state-of-the-art methods. Empirically we compared our method to the recent state-of-the-art IL approaches OPOLO \cite{zhu2020off}, f-IRL \cite{firl2020corl}, and FORM \cite{jaegle2021imitation} in complex continuous control environments. We demonstrate that our method is superior especially if the selection of the best policy cannot be based on the true environment reward signal. This is a setting which more closely resembles real-world applications in autonomous driving or robotics where it is difficult to define a reward function \cite{osa2018an}.  

\section{Background}\label{sec:back}

In this work, we want to train a stochastic policy function $\pi_{\theta}(a_t|s_t)$ in continuous action spaces with parameters $\theta$ in a sequential decision making task considering finite-horizon environments\footnote{An extension to infinite-horizons is given in section \ref{sec:method}}. 
The problem is modeled as a Markov Decision Process (MDP), which is described by the tuple $(S, A, p, r)$ with the continuous state spaces $S$ and action spaces $A$. The transition probability is described by $p(s_{t+1}|s_t, a_t)$ and the bounded reward function by $r(s_t, a_t)$. At every time step $t$ the agent interacts with its environment by observing a state $s_t$ and taking an action $a_t$. This results in a new state $s_{t+1}$ and a reward signal $r_{t+1}$ based on the transition probability and reward function. We will use $\mu^{\pi_{\theta}}(s_t,a_t)$ to denote the state-action marginals at time step t of the trajectory distribution induced by the policy $\pi_{\theta}(a_t|s_t)$.  

\subsection{Maximum Entropy Reinforcement Learning and Soft Actor Critic}\label{sec:sac}

The standard objective in RL is the expected sum of undiscounted rewards $\sum_{t=0}^{T}\mathbb{E}_{(s_t,a_t) \sim \mu^{\pi_{\theta}}}[r(s_t, a_t)]$. The goal of the agent is to learn a policy $\pi_{\theta}(a_t|s_t)$ which maximises this objective. The maximum entropy objective 
\begin{equation}
    J(\pi_{\theta})=\sum_{t=0}^{T}\mathbb{E}_{(s_t,a_t) \sim \mu^{\pi_{\theta}}}[r(s_t, a_t)+\alpha \mathcal{H}(\pi_{\theta}(\cdot |s))]
\end{equation}
introduces a modified goal for the RL agent, where the agent has to maximise the sum of the reward signal and its output entropy $\mathcal{H}(\pi_{\theta}(\cdot |s))$ \cite{Ziebart2010}. The parameter $\alpha$ controls the stochasticity of the optimal policy by determining the relative importance of the entropy term versus the reward. 

Soft Actor-Critic (SAC) \cite{haarnoja2018SAC,haarnoja2018sacapps} combines off-policy Q-Learning with a stable stochastic actor-critic formulation. 
The soft Q-function parameters $\Psi$ can be trained with:
\begin{equation}\label{qloss}
\begin{aligned}
J_Q  = \mathbb{E}_{(s_t,a_t) \sim D_{RB}}[
\frac{1}{2}(Q_{\Psi}(s_t,a_t) -  (r(s_t,a_t) & \\ + \gamma \mathbb{E}_{s_{t+1}}[V_{\hat{\Psi}}(s_{t+1})]))^2]
\end{aligned}
\end{equation}
The soft state value function is defined by:
\begin{equation}
V^{\pi}(s_t):=\mathbb{E}_{a_t \sim \pi}[Q(s_t,a_t) - \alpha \log \pi_{\theta}(a_t|s_t)]
\end{equation}
 Lastly, the policy is optimized by minimizing the following objective:
\begin{equation}\label{piloss}
J_{\pi} = \mathbb{E}_{(s_t) \sim D_{RB}}[
\mathbb{E}_{(a_t) \sim \pi_{\theta}}[\alpha \log \pi_{\theta}(a_t|s_t) - Q_{\Psi}(s_t,a_t)]
]
\end{equation}

\subsection{Imitation Learning}

In the IL setup, the agent does not have access to the true environment reward function $r(s_t, a_t)$ and instead has to imitate expert trajectories performed by an expert policy $\pi_{E}$ collected in a dataset $\mathcal{D}_E$. 

In the typical \textbf{learning-from-demonstration} setup the expert demonstrations $\mathcal{D}^{LfD}_E:=\{s^k_t,a^k_t, s^k_{t+1}\}^N_{k=1}$ are given by action-state-next-state transitions. Distribution matching has been a popular choice among different LfD approaches. The policy $\pi_{\theta}$ is learned by minimizing the discrepancy between the stationary state-action distribution induced by the expert $\mu^{E}(s,a)$ and the policy $\mu^{\pi_{\theta}}(s,a)$. An overview and comparison of different LfD objectives resulting from this discrepancy minimization was done by Ghasemipour et al. \shortcite{ghasemipour19}. Often the backward KLD is used to measure this discrepancy \cite{fu2017learning}: 

\begin{equation}
\min J_{LfD}(\pi_{\theta}) := \min \mathbb{D}_{KL}(\mu^{\pi_{\theta}}(s,a)||\mu^{E}(s,a))    
\end{equation}

\textbf{Learning-from-observation} (LfO) considers a more challenging task where expert actions are not available. Hence, the demonstrations $\mathcal{D}^{LfO}_E:=\{s^k_t, s^k_{t+1}\}^N_{k=1}$ consist of state-next-state transitions.
The policy learns which actions to take based on interactions with the environment and the expert state transitions. 
Distribution matching based on state-transition distributions is a popular choice for state-only IL \cite{torabi2018generative,zhu2020off}: 

\begin{equation}
\min J_{LfO}(\pi_{\theta}) := \min \mathbb{D}_{KL}(\mu^{\pi_{\theta}}(s,s')||\mu^{E}(s,s'))    
\end{equation}

\section{Method}\label{sec:method}
In a finite horizon MDP setting, the joint state-only trajectory distributions are defined by the start state distribution $p(s_0)$ and the product of the conditional state transition distributions $p(s_{i+1} | s_i)$. For the policy distribution $\mu^{\pi_{\theta}}$ and the expert distribution $\mu^{E}$ this becomes:
\[\mu^{\pi_{\theta}}(s_T,..,s_0) = p(s_0) \prod_{i = 0..T-1} \mu^{\pi_{\theta}}(s_{i+1}|s_i),\]

\[\mu^{E}(s_T,..,s_0) = p(s_0)  \prod_{i = 0..T-1} \mu^{E}(s_{i+1}|s_i)\]

Our goal is to match the state-only trajectory distribution $\mu^{\pi_{\theta}}$ induced by the policy with the state-only expert trajectory distribution $\mu^{E}$ by minimizing the Kulback-Leibler divergence (KLD) between them:  

\begin{equation}\label{eq:kldsoiltdm_base}
\begin{aligned}
J_{SOIL-TDM} & = \mathbb{D}_{KL}(\mu^{\pi_{\theta}}||\mu^{E}) \\
& = \mathbb{E}_{(s_T,...,s_0) \sim \mu^{\pi_{\theta}}}[\log \mu^{\pi_{\theta}} - \log \mu^{E}]\\
& = \sum_{i = 0..T-1} \mathbb{E}_{(s_{i+1},s_i) \sim \mu^{\pi_{\theta}}}[\log  \mu^{\pi_{\theta}}(s_{i+1}|s_i) \\ & \hspace{2.0cm} - \log  \mu^{E}(s_{i+1}|s_i)]
\end{aligned}
\end{equation}

The conditional expert state transition distribution $ \mu^{E}(s_{i+1}|s_{i}) $ can be learned offline from the demonstrations for example by training a conditional normalizing flow on the given state/next-state pairs. The policy induced conditional state transition distribution can be rewritten with the Bayes theorem using the environment model $p(s_{i+1}|a_i,s_i)$ and the inverse action distribution density $\pi'_{\theta}(a_{i}|s_{i+1},s_i)$:
\begin{equation}\label{eq:seperation}
\mu^{\pi_{\theta}}(s_{i+1}|s_i) = \frac{p(s_{i+1}|a_i,s_i) \pi_{\theta}(a_i|s_i)}{\pi'_{\theta}(a_{i}|s_{i+1},s_i)}
\end{equation}
It holds for any $a_i$ where $\pi' > 0$ . Thus, one can extend the expectation over $(s_{i+1},s_i)$ by the action $a_i$ and the KLD minimization $\min \mathbb{D}_{KL}(\mu^{\pi_{\theta}}||\mu^{E})$ can be rewritten as
\begin{equation}\label{eq:kldsoiltdm}
\begin{aligned}
\min \sum_{i = 0..T-1} & \mathbb{E}_{(s_i,a_i,s_{i+1}) \sim {\pi_{\theta}}}[\log p(s_{i+1}|a_i,s_i) + \log \pi_{\theta}(a_i|s_i) \\ & - \log \pi'_{\theta}(a_{i}|s_{i+1},s_i)  - \log  \mu^{E}(s_{i+1}|s_i)]
\end{aligned}
\end{equation}
Now, by defining a reward function (also see \ref{appendix:assumptions})
\begin{equation}\label{eq:rewardsoiltdm}
\begin{aligned}
r(a_i, s_i) := & \mathbb{E}_{s_{i+1} \sim p(s_{i+1}|a_i,s_i)} [-\log p(s_{i+1}|a_i,s_i) \\&  + \log \pi'_{\theta}(a_{i}|s_{i+1},s_i) + \log  \mu^{E}(s_{i+1}|s_i)]
\end{aligned}
\end{equation}
 that depends on the expert state transition likelihood $\mu^{E}(s_{i+1}|s_i)$, on the environment model $p(s_{i+1}|a_i,s_i)$ and on the inverse action distribution density $\pi'_{\theta}(a_{i}|s_{i+1},s_i)$. The state-only trajectory distribution matching problem can be transformed to a max-entropy RL task:
\begin{equation}\label{eq:maxentropyequivalence}
\begin{aligned}
& \min \mathbb{D}_{KL}(\mu^{\pi_{\theta}}||\mu^{E})  \\& = \max \sum_{i = 0..T-1}  \mathbb{E}_{(a_i,s_i) \sim \pi_{\theta}}[ -\log \pi_{\theta}(a_i|s_i) + r(a_i, s_i)]\\
& = \max \sum_{i = 0..T-1}  \mathbb{E}_{(a_i,s_i) \sim \pi_{\theta}}[r(a_i, s_i) + \mathcal{H}(\pi_{\theta}(\cdot|s)]
\end{aligned}
\end{equation}
In practice the reward function $r(a_i,s_i)$ can be computed using monte carlo integration with a single sample from $p(s_{i+1}|a_i,s_i)$ using the replay buffer. 

This max-entropy RL task can be optimized with standard max-entropy RL algorithms. In this work, we applied the SAC algorithm \cite{haarnoja2018SAC} as it is outlined in Section \ref{sec:sac}. 

The extension to infinite horizon tasks can be done by introducing a discount factor $\gamma$ as in the work by Haarnoja et al. \shortcite{haarnoja2018sacapps}. In combination with our reward definition one obtains the following infinite horizon maximum entropy objective:
\begin{equation}
\begin{aligned}
J_{ME-iH} = & \sum_{i = 0..\inf}  \mathbb{E}_{(a_i,s_i) \sim \pi_{\theta}}[ \sum_{j = i..\inf} \gamma^{j-i} \\& \mathbb{E}_{(a_j,s_j) \sim \pi_{\theta}} [r(a_j, s_j) + \mathcal{H}(\pi_{\theta}(\cdot|s_j) | s_i, a_i] ]
\end{aligned}
\end{equation}

\subsection{Algorithm}
To evaluate the reward function, the environment model $p(s_{i+1}|a_i,s_i)$ and the inverse action distribution function $\pi'_{\theta}(a_{i}|s_{i+1},s_i)$ have to be estimated. We model both distributions using conditional normalizing flows and train with maximum likelihood based on expert demonstrations and rollout data from a replay buffer. The environment model $p(s_{i+1}|a_i,s_i)$ is modeled by $\mu_{\phi}(s_{i+1}|a_i,s_i)$ with parameter $\phi$ and the inverse action distribution function $\pi'_{\theta}(a_{i}|s_{i+1},s_i)$ is modeled by $\mu_{\eta}(a_{i}|s_{i+1},s_i)$ with parameter $\eta$. 

The whole training process according to Algorithm \ref{alg:SOIL-TDM} is described in the following\footnote{Code will be available at https://github.com/*}. The expert state transition model $\mu^E(s_{t+1}|s_t)$ is trained offline using the expert dataset $D_E$ which contains $K$ expert state trajectories. We assume the expert distribution is correctly represented by the dataset. Density modeling of the expert state transitions can still result in overfitting when only few expert demonstrations are available (in the few sample limit). We improved the expert training process by adding Gaussian noise to the state values. The standard deviation of the noise is reduced during training so that the model has a correct estimate of the density without overfitting to the explicit demonstrations. With this approach we are able to successfully train expert state transition models on as few as one expert trajectory. The influence of this improved routine is studied in Appendix \ref{sec:ablStud}. After this initial step the following process is repeated until convergence in each episode. The policy interacts with the environment for $T$ steps to collect state-action-next-state information, which is saved in the replay buffer $D_{RB}$. The conditional normalizing flows for the environment model $\mu_{\phi}(s_{i+1}|a_i,s_i)$ (policy independent) and the inverse action distribution model $\mu_{\eta}(a_{i}|s_{i+1},s_i)$ (policy dependent) are optimized using samples from the replay buffer $D_{RB}$ for $N$ steps. In Appendix \ref{appendix:replaybuffer} we show that this reduces the KLD (Equation \ref{eq:kldsoiltdm}) in each step. Afterwards we use the learned models together with the samples from the replay buffer to compute a one-sample monte carlo approximation of the reward to train the Q-function. The policy $\pi_{\theta}(a_t|s_t)$ is updated using SAC. The SAC-based Q-function training and policy optimization also minimize Equation \ref{eq:kldsoiltdm} (see Equation \ref{eq:maxentropyequivalence}) in each step. Together with the inverse action policy learning they lead to a converging algorithm since all steps reduce the KLD, which is bounded from below by $0$. It is worth noting that the overall algorithm is non-adversarial, the inverse action policy optimization and the policy optimization using SAC both reduce the overall objective - the KLD. Contrary to AIL algorithms like OPOLO, it is not based on an adversarial nested min-max optimization. Additionally, we can estimate the similarity of state transitions from our policy to the expert during each optimization step, since we model all densities in the rewritten KLD from Equation \ref{eq:kldsoiltdm}. As a result we have a reliable performance estimate enabling us to select the best performing policy based on the lowest KLD between policy and expert state transition trajectories.

\subsection{Relation to learning from observations}
The LfO objective of previous approaches like OPOLO minimizes the divergence between the joint policy state transition distribution and the joint expert state transition distribution:
\begin{equation}
    J_{LfO} = \mathbb{D}_{KL}(\mu^{\pi_{\theta}}(s',s)||\mu^E(s',s))
\end{equation}
which can be rewritten as (see \ref{appendix:lfo})
\begin{equation}
\begin{aligned}
J_{LfO} = & \mathbb{D}_{KL}(\mu^{\pi_{\theta}}(s_T,...,s_0)||\mu^E(s_T,...,s_0)) 
\\& + \sum_{i = 1..T-1} \mathbb{D}_{KL}(\mu^{\pi_{\theta}}(s_i)||\mu^E(s_i))
\end{aligned}
\end{equation}

Thus, this LfO objective minimizes the sum of the KLD between the joint distributions and the KLDs of the marginal distributions. The SOIL-TDM objective in comparison minimizes purely the KLD of the joint distributions. In case of a perfect distribution matching - a zero KLD between the joint distributions - the KLDs of the marginals also vanish so both objectives have the same optimum.

\begin{algorithm}[t]
\begin{algorithmic}[1]  
\Procedure{SOIL-TDM}{$D_E$}
\State Train $\mu^{E}(s_{t+1}|s_{t})$ \texttt{with} $D_E:\lbrace s_0, s_1, ...s_T \rbrace^K_{k=0}$
\For{\texttt{episodes}}
\For{\texttt{range(T)}} \Comment{generate data}
    \State $\hat{a}_t$ $\gets$ sample($\pi_{\theta}(\hat{a}_t|s_t)$) 
    \State $s_{t+1}$ $\gets$ $p_{sim}(s_{t+1} | s_{t}, \hat{a}_t)$ \Comment{apply action}
    \State \texttt{Store} $(s_t, \hat{a}_t, s_{t+1})$ in $D_{RB}$
\EndFor 
\For{\texttt{range(N)}} \Comment{update dynamics models}
    \State $\{(s_t, \hat{a}_t, s_{t+1})\}^{B}_{i=1} \sim D_{RB}$ \Comment{sample batch}
    \State \texttt{train} $\mu_{\eta}(\hat{a}_t|s_{t+1},s_t)$ and $\mu_{\phi}(s_{t+1}|\hat{a}_t,s_t)$ 
\EndFor
\For{\texttt{range(N)}} \Comment{SAC Optimization}
    \State  $\{(s_t, \hat{a}_t, s_{t+1})\}^{B}_{i=1} \sim D_{RB}$ \Comment{sample batch}
    \State $a_t$ $\gets$ sample($\pi_{\theta}(a_t|s_t)$)
    \State \texttt{optimize} $\pi_{\theta}(a_t|s_t)$ with $J_{\pi}$ from eq. \ref{piloss} 
    \State \Comment{estimate reward}
    \State $r(s_t,\hat{a}_t)$ $\gets$ $-\log \mu_{\phi}(s_{t+1}|\hat{a}_t,s_t) + \log \mu_{\eta}(\hat{a}_t|s_{t+1},s_t) + \log  \mu^{E}(s_{t+1}|s_t)$ 
    \State \texttt{optimize} $Q_{\psi}(\hat{a}_t,s_t)$ with $J_{Q}$ from eq. \ref{qloss}  
\EndFor
\EndFor
\EndProcedure  
\end{algorithmic}  
\caption{State-Only Imitation Learning by Trajectory Distribution Matching (SOIL-TDM)}\label{alg:SOIL-TDM}
\end{algorithm}

\section{Related Work}

Many recent IL approaches are based on inverse RL (IRL) \cite{Ng2000AlgorithmsFI}. In IRL, the goal is to learn a reward signal for which the expert policy is optimal. AIL algorithms are popular methods to perform IL in a RL setup \cite{GAIL,fu2017learning,Kostrikov2020Imitation}. In AIL, a discriminator gets trained to distinguish between expert states and states coming from policy rollouts. The goal of the policy is to fool the discriminator. The policy gets optimized to match the state action distribution of the expert, using this two-player game. Based on this idea more general approaches have been derived based on f-divergences. Ni et al.\shortcite{firl2020corl} derive an analytic gradient of any f-divergence between the agent and expert state distribution w.r.t. reward parameters. Based on this gradient they present the algorithm f-IRL that recovers a stationary reward function from the expert density by gradient descent. Ghasemipour et al.\shortcite{ghasemipour19} identify that IRL’s state-marginal matching objective contributes most to its superior performance and apply this understanding to teach agents a diverse range of behaviours using simply hand-specified state distributions.

A key problem with AIL for LfD and LfO is optimization instability \cite{miyato2018spectral}. Wang et al.\shortcite{wang2019random} avoid the instabilities resulting from adversarial optimization by estimating the support of the expert policy to compute a fixed reward function. Similarly, Brantley et al.\shortcite{brantley2020disagreement} use a fixed reward function by estimating the variance of an ensemble of policies. Both methods rely on additional behavioral cloning steps to reach expert-level performance. Liu et  al.\shortcite{liu2020energy} propose Energy-Based Imitation Learning (EBIL) which recovers fixed and interpretative reward signals by directly estimating the expert’s energy. Neural Density Imitation (NDI) \cite{kim2021imitation} uses density models to perform distribution matching. Deterministic and Discriminative Imitation (D2-Imitation) \cite{sun2021deterministic} requires no adversarial training by partitioning samples into two replay buffers and then learning a deterministic policy via off-policy reinforcement learning. Inverse soft-Q learning (IQ-Learn) \cite{garg2021iqlearn} avoids adversarial training by learning a single Q-function to implicitly representing both reward and policy. The implicitly learned rewards from IQ-Learn show a high positive correlation with the ground-truth rewards.  

LfO can be divided into model-free and model-based approaches. GAILfO \cite{torabi2018generative} is a model-free approach which uses the GAIL principle with the discriminator input being state-only. Yang et al. \shortcite{yang2019imitation} analyzed the gap between the LfD and LfO objectives and proved that it lies in the disagreement of inverse dynamics models between the imitator and expert. Their proposed method IDDM is based on an upper bound of this gap in a model-free way. OPOLO \cite{zhu2020off} is a sample-efficient LfO approach also based on AIL, which enables off-policy optimization. The policy update is also regulated with an inverse action model that assists distribution matching in a mode-covering perspective. 

Other model-based approaches either apply forward dynamics models or inverse action models. Sun et al. \shortcite{sun2019provably} proposed a solution based on forward dynamics models to learn time dependant policies. While being provably efficient, it is not suited for infinite horizon tasks. Alternatively, behavior cloning from observations (BCO) \cite{torabi2018behavioral} learns an inverse action model based on simulator interactions to infer actions based on the expert state demonstrations. GPRIL \cite{schroecker2019generative} uses normalizing flows as generative models to learn backward dynamics models to estimate predecessor transitions and augmenting the expert data set with further trajectories, which lead to expert states.  Jiang et al. \shortcite{jiang2020offline} investigated IL using few expert demonstrations and a simulator with misspecified dynamics. A detailed overview of LfO was done by Torabi et al. \shortcite{torabi2019recent}.

\subsection{Method Discussion and Relation to FORM}
While our proposed method SOIL-TDM was independently developed it is most similar to the state-only approach FORM \cite{jaegle2021imitation}. In FORM the policy training is guided by a conditional density estimation of the expert’s observed state transitions. In addition a state transition model $\mu^{\pi_{\theta}}_{\Phi}(s_{i+1}|s_{i})$ of the current policy is learned. The policy reward is estimated by: $ r_i = log \mu^E(s_{i+1}|s_{i}) - log \mu^{\pi_{\theta}}_{\Phi}(s_{i+1}|s_{i})$. The approach matches conditional state transition probabilities of expert and policy in comparison to the joint state-action (like GAIL) or joint state-next-state (like OPOLO or GAILfO) densities. The authors argue that this has consequences in the robustness of the different approaches. Namely, methods  based on conditional state probabilities are less sensitive to erroneously penalizing features that may not be in the demonstrator data but lead to correct transitions. Hence, such methods may be less prone to overfit to irrelevant differences. Jaegle et al. \shortcite{jaegle2021imitation} demonstrate the benefit of such a conditional density matching approach. 

In contrast to FORM we show in Equation \ref{eq:seperation} that the policies next-state conditional density $\mu^{\pi_{\theta}}(s_{i+1}|s_{i})$ can be separated into the policies action density and the forward- and the backward-dynamics densities. Using this decomposition we show that the KLD minimization is equivalent to a maximum entropy RL objective (see Equation \ref{eq:kldsoiltdm}) with a special reward (see Equation \ref{eq:rewardsoiltdm}). Here the entropy of the policy stemming from the decomposition of the conditional state-next-state density leads to the maximum entropy RL objective. Jaegle et al. \shortcite{jaegle2021imitation} mention that the second term in their reward objective can be viewed as an entropy-like expression. Hence, if this reward is optimized using a RL algorithm which includes some form of policy entropy regularization this entropy is basically weighted twice. In the experiments we show that this double accounting of the policy entropy negatively affects the sample efficiency of the algorithm in comparison to our method.

\section{Experiments}
We evaluate our proposed method described in Section \ref{sec:method} in a variety of different IL tasks and compare it against the baseline methods OPOLO, F-IRL and FORM. We evaluate and compare all methods in complex and high dimensional continuous control environments using the Pybullet physics simulation \cite{coumans2019}. To evaluate the performance of all methods, the episode rewards of the trained policies are compared to reward from the expert policy. The expert data generation as well as the used baseline implementations are described in Appendix \ref{appendix:expSetup}. 

Since we assume no environment reward is available as an early stopping criterion, we use other convergence estimates available during training to select the best policy for each method. In adversarial training the duality gap \cite{grnarova2019a,Sidheekh2021OnDG} is an established method to estimate the convergence of the training process. In the IL setup the duality gap can be very difficult to estimate since it requires the gradient of the policy and the environment (i.e. the gradient of the generator) for the optimization process it relies on. We therefore use two alternatives for model selection for OPOLO. The first approach selects the model with the lowest policy loss and the second approach selects the model based on the highest estimated reward over ten consecutive epochs. For F-IRL we selected the model with the lowest estimated Jensen-Shannon divergence over ten epochs. 
To estimate the convergence of SOIL-TDM the policy loss based on the KLD from equation \ref{eq:kldsoiltdm} is used. It can be estimated using the same models used for training the policy. Similarly, we used the effect models of FORM to estimate the convergence based on the reward. 

The evaluation is done by running 3 training runs with ten test episodes (in total 30 rollouts) for each trained policy and calculating the respective mean and confidence interval for all runs. We plot the rewards normalized so that 1 corresponds to expert performance. Values above 1 mean that the agent has achieved a higher reward than the mean of the expert (episode rewards of the expert are reported in Appendix \ref{appendix:expSetup}). Implementation details of our method are described in Appendix \ref{sec:implDet}.

\begin{figure}[t]
\begin{center}
\includegraphics[width=0.5\textwidth]{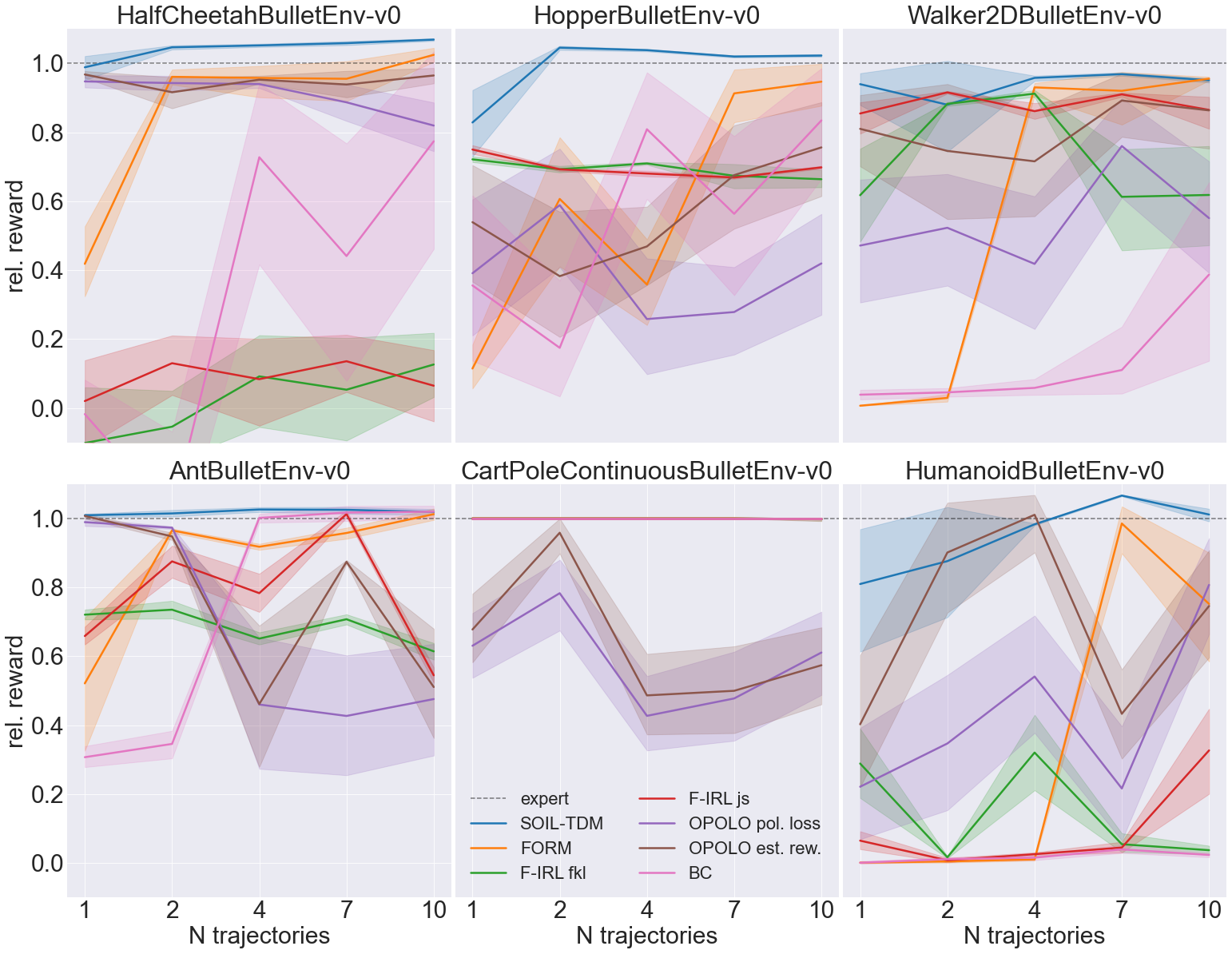}
\end{center}
\caption{Unkown true environment reward selection criteria: Relative reward for different amount of expert trajectories on continuous control environments. The best policies based on estimated convergence values were selected. The value 1 corresponds to expert policy performance.}
\label{fig:reward_plot}
\end{figure}

The evaluation results of the discussed methods on a suite of continuous control tasks with unkown true environment reward as a selection criterion are shown in Figure \ref{fig:reward_plot}. The achieved rewards are plotted with respect to the number of expert trajectories provided for training the agent. The confidence intervals are plotted using lighter colors. 

\begin{figure}[t]
\begin{center}
\includegraphics[width=0.5\textwidth]{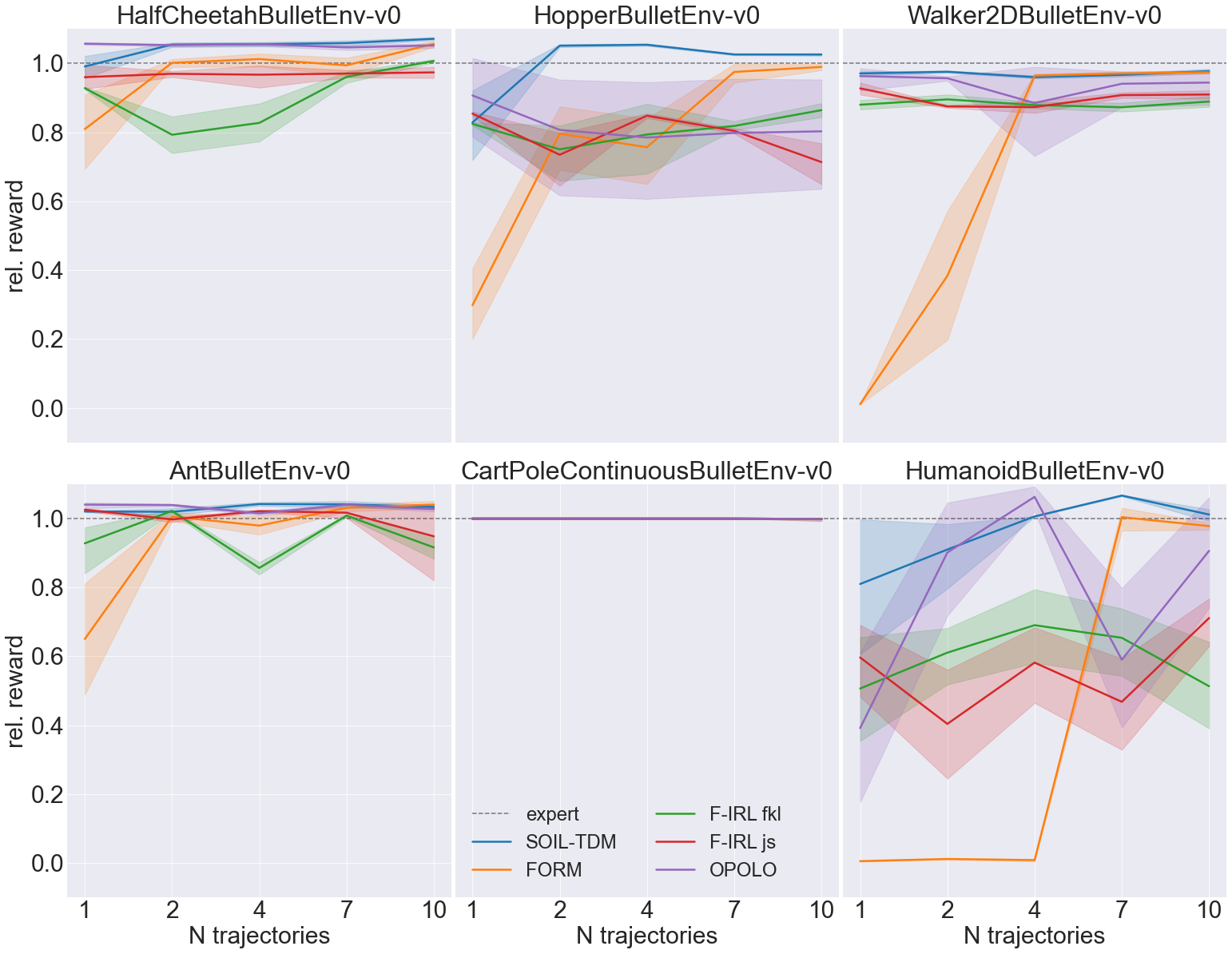}
\end{center}
\caption{Best true environment reward selection criterion: Relative reward for different amount of expert trajectories on continuous control environments. The value 1 corresponds to expert policy performance.}
\label{fig:reward_max_plot}
\end{figure}

If the true environment reward is unkown the results show that SOIL-TDM achieves or surpasses the performance of the baseline methods on all tested environments (with the exception of two and four expert trajectories in the HumanoidBulletEnv-v0 environment and two trajectories in the Walker2DBulletEnv-v0 environment). In general the adversarial based methods OPOLO and F-IRL exhibit a high variance of the achieved rewards using the proposed selection criteria. While loss and reward are well suited for selecting the best model in usual setups, the results demonstrate that they might be less expressive for estimating the convergence in adversarial training due to the min-max game of the discriminator and the policy. The stability of the SOIL-TDM training method is evident from the small confidence band of the results which gets smaller for more expert demonstrations. While the selection of FORM is also more stable compared to the adversarial methods it generally achieves lower rewards in the sample efficient regime of one and two expert trajectories. 

Figure \ref{fig:reward_max_plot} shows the benchmark results of OPOLO, F-IRL, FORM, and SOIL-TDM if the true environment reward is used as an early stopping criterion. In this setup, our method still achieves competitive performance or surpasses OPOLO, F-IRL, and FORM. 
Additional figures for a comparison of the training performance and efficiency can be found in the Appendix \ref{sec:addRes}. An ablation study for our method is in the Appendix \ref{sec:ablStud}. 

\section{Conclusion}
In this work we propose a non-adversarial state-only imitation learning approach based on the minimization of the Kulback-Leibler divergence between the policy and the expert state trajectory distribution. This objective leads to a maximum entropy reinforcement learning problem with a reward function depending on the expert state transition distribution and the forward and backward dynamics of the environment which can be modeled using conditional normalizing flows. The proposed approach is compared to several state-of-the-art learning from observations methods in a scenario with unkown environment rewards and achieves state-of-the-art performance.


\begin{small}
\bibliographystyle{named}
\bibliography{ijcai22}

\begin{thebibliography}{}

\bibitem[\protect\citeauthoryear{Ardizzone \bgroup \em et al.\egroup
  }{2019}]{ardizzone2018analyzing}
Lynton Ardizzone, Jakob Kruse, Carsten Rother, and Ullrich Köthe.
\newblock Analyzing inverse problems with invertible neural networks.
\newblock In {\em International Conference on Learning Representations,
  {ICLR}}, 2019.

\bibitem[\protect\citeauthoryear{{Brantley} \bgroup \em et al.\egroup
  }{2020}]{brantley2020disagreement}
Kiante {Brantley}, Wen {Sun}, and Mikael {Henaff}.
\newblock Disagreement-regularized imitation learning.
\newblock In {\em International Conference on Learning Representations,
  {ICLR}}, 2020.

\bibitem[\protect\citeauthoryear{Coumans and Bai}{2016  2019}]{coumans2019}
Erwin Coumans and Yunfei Bai.
\newblock Pybullet, a python module for physics simulation for games, robotics
  and machine learning.
\newblock \url{http://pybullet.org}, 2016--2019.

\bibitem[\protect\citeauthoryear{Dinh \bgroup \em et al.\egroup
  }{2017}]{realNVP}
Laurent Dinh, Jascha Sohl-Dickstein, and Samy Bengio.
\newblock Density estimation using real nvp, 2017.

\bibitem[\protect\citeauthoryear{{Fu} \bgroup \em et al.\egroup
  }{2017}]{fu2017learning}
Justin {Fu}, Katie {Luo}, and Sergey {Levine}.
\newblock Learning robust rewards with adversarial inverse reinforcement
  learning.
\newblock In {\em International Conference on Learning Representations,
  {ICLR}}, 2017.

\bibitem[\protect\citeauthoryear{Garg \bgroup \em et al.\egroup
  }{2021}]{garg2021iqlearn}
Divyansh Garg, Shuvam Chakraborty, Chris Cundy, Jiaming Song, and Stefano
  Ermon.
\newblock {IQ}-learn: Inverse soft-q learning for imitation.
\newblock In {\em Thirty-Fifth Conference on Neural Information Processing
  Systems}, 2021.

\bibitem[\protect\citeauthoryear{Ghasemipour \bgroup \em et al.\egroup
  }{2019}]{ghasemipour19}
Seyed Kamyar~Seyed Ghasemipour, Richard~S. Zemel, and Shixiang Gu.
\newblock A divergence minimization perspective on imitation learning methods.
\newblock In {\em 3rd Annual Conference on Robot Learning, CoRL}, Proceedings
  of Machine Learning Research, 2019.

\bibitem[\protect\citeauthoryear{{Grnarova} \bgroup \em et al.\egroup
  }{2019}]{grnarova2019a}
Paulina {Grnarova}, Kfir~Y. {Levy}, Aurelien {Lucchi}, Nathanael {Perraudin},
  Ian {Goodfellow}, Thomas {Hofmann}, and Andreas {Krause}.
\newblock A domain agnostic measure for monitoring and evaluating gans.
\newblock In {\em Advances in Neural Information Processing Systems}, 2019.

\bibitem[\protect\citeauthoryear{Haarnoja \bgroup \em et al.\egroup
  }{2018a}]{haarnoja2018SAC}
Tuomas Haarnoja, Aurick Zhou, Pieter Abbeel, and Sergey Levine.
\newblock Soft actor-critic: Off-policy maximum entropy deep reinforcement
  learning with a stochastic actor.
\newblock In {\em Proceedings of the 35th International Conference on Machine
  Learning}, 2018.

\bibitem[\protect\citeauthoryear{Haarnoja \bgroup \em et al.\egroup
  }{2018b}]{haarnoja2018sacapps}
Tuomas Haarnoja, Aurick Zhou, Kristian Hartikainen, George Tucker, Sehoon Ha,
  Jie Tan, Vikash Kumar, Henry Zhu, Abhishek Gupta, Pieter Abbeel, and Sergey
  Levine.
\newblock Soft actor-critic algorithms and applications.
\newblock Technical report, 2018.

\bibitem[\protect\citeauthoryear{Ho and Ermon}{2016}]{GAIL}
Jonathan Ho and Stefano Ermon.
\newblock Generative adversarial imitation learning.
\newblock In {\em Advances in Neural Information Processing Systems}. Curran
  Associates, Inc., 2016.

\bibitem[\protect\citeauthoryear{Jaegle \bgroup \em et al.\egroup
  }{2021}]{jaegle2021imitation}
Andrew Jaegle, Yury Sulsky, Arun Ahuja, Jake Bruce, Rob Fergus, and Greg Wayne.
\newblock Imitation by predicting observations.
\newblock In {\em Proceedings of the 38th International Conference on Machine
  Learning}, 2021.

\bibitem[\protect\citeauthoryear{{Jiang} \bgroup \em et al.\egroup
  }{2020}]{jiang2020offline}
Shengyi {Jiang}, Jingcheng {Pang}, and Yang {Yu}.
\newblock Offline imitation learning with a misspecified simulator.
\newblock In {\em Advances in Neural Information Processing Systems}, 2020.

\bibitem[\protect\citeauthoryear{Jin \bgroup \em et al.\egroup
  }{2020}]{Jin2020What}
Chi Jin, Praneeth Netrapalli, and Michael Jordan.
\newblock What is local optimality in nonconvex-nonconcave minimax
  optimization?
\newblock In {\em Proceedings of the 37th International Conference on Machine
  Learning}, 2020.

\bibitem[\protect\citeauthoryear{{Kim} \bgroup \em et al.\egroup
  }{2021}]{kim2021imitation}
Kuno {Kim}, Akshat {Jindal}, Yang {Song}, Jiaming {Song}, Yanan {Sui}, and
  Stefano {Ermon}.
\newblock Imitation with neural density models.
\newblock In {\em arxiv:cs.LG}, 2021.

\bibitem[\protect\citeauthoryear{Kingma and Ba}{2015}]{KingmaB14}
Diederik~P. Kingma and Jimmy Ba.
\newblock Adam: {A} method for stochastic optimization.
\newblock In {\em International Conference on Learning Representations,
  {ICLR}}, 2015.

\bibitem[\protect\citeauthoryear{Kingma and
  Dhariwal}{2018}]{NEURIPS2018_d139db6a}
Durk~P Kingma and Prafulla Dhariwal.
\newblock Glow: Generative flow with invertible 1x1 convolutions.
\newblock In {\em Advances in Neural Information Processing Systems}, 2018.

\bibitem[\protect\citeauthoryear{Kostrikov \bgroup \em et al.\egroup
  }{2020}]{Kostrikov2020Imitation}
Ilya Kostrikov, Ofir Nachum, and Jonathan Tompson.
\newblock Imitation learning via off-policy distribution matching.
\newblock In {\em International Conference on Learning Representations,
  {ICLR}}, 2020.

\bibitem[\protect\citeauthoryear{{Kuefler} \bgroup \em et al.\egroup
  }{2017}]{kuefler2017imitating}
Alex {Kuefler}, Jeremy {Morton}, Tim {Wheeler}, and Mykel {Kochenderfer}.
\newblock Imitating driver behavior with generative adversarial networks.
\newblock In {\em 2017 IEEE Intelligent Vehicles Symposium (IV)}, 2017.

\bibitem[\protect\citeauthoryear{{Liu} \bgroup \em et al.\egroup
  }{2020}]{liu2020energy}
Minghuan {Liu}, Tairan {He}, Minkai {Xu}, and Weinan {Zhang}.
\newblock Energy-based imitation learning.
\newblock {\em Autonomous Agents and Multi-Agent Systems}, 2020.

\bibitem[\protect\citeauthoryear{{Miyato} \bgroup \em et al.\egroup
  }{2018}]{miyato2018spectral}
Takeru {Miyato}, Toshiki {Kataoka}, Masanori {Koyama}, and Yuichi {Yoshida}.
\newblock Spectral normalization for generative adversarial networks.
\newblock In {\em International Conference on Learning Representations,
  {ICLR}}, 2018.

\bibitem[\protect\citeauthoryear{Ng and Russell}{2000}]{Ng2000AlgorithmsFI}
A.~Ng and S.~Russell.
\newblock Algorithms for inverse reinforcement learning.
\newblock In {\em "International Conference on Machine Learning"}, 2000.

\bibitem[\protect\citeauthoryear{Ni \bgroup \em et al.\egroup
  }{2020}]{firl2020corl}
Tianwei Ni, Harshit Sikchi, Yufei Wang, Tejus Gupta, Lisa Lee, and Ben
  Eysenbach.
\newblock f-irl: Inverse reinforcement learning via state marginal matching.
\newblock In {\em Conference on Robot Learning}, 2020.

\bibitem[\protect\citeauthoryear{{Osa} \bgroup \em et al.\egroup
  }{2018}]{osa2018an}
Takayuki {Osa}, Joni {Pajarinen}, Gerhard {Neumann}, J.~Andrew {Bagnell},
  Pieter {Abbeel}, and Jan {Peters}.
\newblock An algorithmic perspective on imitation learning.
\newblock In {\em Foundations and Trends in Robotics}, 2018.

\bibitem[\protect\citeauthoryear{{Papamakarios} \bgroup \em et al.\egroup
  }{2019}]{papamakarios2019normalizing}
George {Papamakarios}, Eric~T. {Nalisnick}, Danilo~Jimenez {Rezende}, Shakir
  {Mohamed}, and Balaji {Lakshminarayanan}.
\newblock Normalizing flows for probabilistic modeling and inference.
\newblock {\em Journal of Machine Learning Research}, 2019.

\bibitem[\protect\citeauthoryear{{Schroecker} \bgroup \em et al.\egroup
  }{2019}]{schroecker2019generative}
Yannick {Schroecker}, Mel {Vecerík}, and Jonathan {Scholz}.
\newblock Generative predecessor models for sample-efficient imitation
  learning.
\newblock In {\em International Conference on Learning Representations,
  {ICLR}}, 2019.

\bibitem[\protect\citeauthoryear{Sidheekh \bgroup \em et al.\egroup
  }{2021}]{Sidheekh2021OnDG}
Sahil Sidheekh, Aroof Aimen, Vineet Madan, and N.~C. Krishnan.
\newblock On duality gap as a measure for monitoring gan training.
\newblock {\em 2021 International Joint Conference on Neural Networks (IJCNN)},
  2021.

\bibitem[\protect\citeauthoryear{{Sun} \bgroup \em et al.\egroup
  }{2019}]{sun2019provably}
Wen {Sun}, Anirudh {Vemula}, Byron {Boots}, and Drew {Bagnell}.
\newblock Provably efficient imitation learning from observation alone.
\newblock In {\em Proceedings of the 36th International Conference on Machine
  Learning}, 2019.

\bibitem[\protect\citeauthoryear{Sun \bgroup \em et al.\egroup
  }{2021}]{sun2021deterministic}
Mingfei Sun, Sam Devlin, Katja Hofmann, and Shimon Whiteson.
\newblock Deterministic and discriminative imitation (d2-imitation): Revisiting
  adversarial imitation for sample efficiency.
\newblock In {\em Association for the Advancement of Artificial Intelligence},
  2021.

\bibitem[\protect\citeauthoryear{{Torabi} \bgroup \em et al.\egroup
  }{2018a}]{torabi2018behavioral}
Faraz {Torabi}, Garrett {Warnell}, and Peter {Stone}.
\newblock Behavioral cloning from observation.
\newblock In {\em Proceedings of the Twenty-Seventh International Joint
  Conference on Artificial Intelligence}, 2018.

\bibitem[\protect\citeauthoryear{{Torabi} \bgroup \em et al.\egroup
  }{2018b}]{torabi2018generative}
Faraz {Torabi}, Garrett {Warnell}, and Peter {Stone}.
\newblock Generative adversarial imitation from observation.
\newblock In {\em International Conference on Machine Learning Workshop on
  Imitation, Intent, and Interaction (I3)}, 2018.

\bibitem[\protect\citeauthoryear{{Torabi} \bgroup \em et al.\egroup
  }{2019}]{torabi2019recent}
Faraz {Torabi}, Garrett {Warnell}, and Peter {Stone}.
\newblock Recent advances in imitation learning from observation.
\newblock In {\em Proceedings of the Twenty-Eighth International Joint
  Conference on Artificial Intelligence}, 2019.

\bibitem[\protect\citeauthoryear{{Wang} \bgroup \em et al.\egroup
  }{2019}]{wang2019random}
Ruohan {Wang}, Carlo {Ciliberto}, Pierluigi~Vito {Amadori}, and Yiannis
  {Demiris}.
\newblock Random expert distillation: Imitation learning via expert policy
  support estimation.
\newblock In {\em Proceedings of the 36th International Conference on Machine
  Learning}, 2019.

\bibitem[\protect\citeauthoryear{{Yang} \bgroup \em et al.\egroup
  }{2019}]{yang2019imitation}
Chao {Yang}, Xiaojian {Ma}, Wenbing {Huang}, Fuchun {Sun}, Huaping {Liu},
  Junzhou {Huang}, and Chuang {Gan}.
\newblock Imitation learning from observations by minimizing inverse dynamics
  disagreement.
\newblock In {\em Advances in Neural Information Processing Systems}, 2019.

\bibitem[\protect\citeauthoryear{{Zhu} \bgroup \em et al.\egroup
  }{2020}]{zhu2020off}
Zhuangdi {Zhu}, Kaixiang {Lin}, Bo~{Dai}, and Jiayu {Zhou}.
\newblock Off-policy imitation learning from observations.
\newblock In {\em Advances in Neural Information Processing Systems}, 2020.

\bibitem[\protect\citeauthoryear{Ziebart}{2010}]{Ziebart2010}
Brian~D. Ziebart.
\newblock {\em Modeling Purposeful Adaptive Behavior with the Principle of
  Maximum Causal Entropy}.
\newblock PhD thesis, 2010.

\end{thebibliography}
\end{small}

\clearpage

\appendix
\section{Appendix}

\subsection{Relation to LfO}\label{appendix:lfo}
The learning from observations (LfO) objective minimizes the divergence between the joint policy state transition distribution and the joint expert state transition distribution:
\begin{equation}
    \min J_{LfO} = \min \mathbb{D}_{KL}(\mu^{\pi_{\theta}}(s',s)||\mu^E(s',s))
\end{equation}
where $s'$ is a successor state of $s$ given a stationary policy and stationary $s',s$ marginals. This can be rewritten as

\begin{equation}
\begin{aligned}
    J_{LfO} & =  \sum_{i = 0..T-1} \mathbb{D}_{KL}(\mu^{\pi_{\theta}}(s_{i+1},s_i)||\mu^E(s_{i+1},s_i))\\
    & = \sum_{i = 0..T-1} \int \mu^{\pi_{\theta}}(s_{i+1},s_i)(\log \mu^{\pi_{\theta}}(s_{i+1},s_i) \\& \hspace{1.0cm} - \log \mu^E(s_{i+1},s_i)) \\
    & =\sum_{i = 0..T-1} \int \mu^{\pi_{\theta}}(s_T,..,s_0)(\log \mu^{\pi_{\theta}}(s_{i+1},s_i) \\&\hspace{1.0cm} - \log \mu^E(s_{i+1},s_i)) \\
    & = \int \mu^{\pi_{\theta}}(s_T,..,s_0) \sum_{i = 0..T-1} (\log \mu^{\pi_{\theta}}(s_{i+1},s_i) \\&\hspace{1.0cm} - \log \mu^E(s_{i+1},s_i)) \\
    & = \int \mu^{\pi_{\theta}}(s_T,..,s_0) \sum_{i = 0..T-1} (\log \mu^{\pi_{\theta}}(s_{i+1}|s_i) \\&\hspace{1.0cm} + \log \mu^{\pi_{\theta}}(s_i) - \log \mu^E(s_{i+1}|s_i)) - \log \mu^E(s_i) \\
    & = \mathbb{E}_{(s_T,...,s_0) \sim \mu^{\pi_{\theta}}}[ \log  \frac{\mu^{\pi_{\theta}}(s_T,...,s_0)}{\mu^E(s_T,...,s_0)} \\&\hspace{1.0cm} + \log \prod_{i = 1..T-1} \frac{\mu^{\pi_{\theta}}(s_i)}{\mu^E(s_i)} ] \\
    & = \mathbb{D}_{KL}(\mu^{\pi_{\theta}}(s_T,...,s_0)||\mu^E(s_T,...,s_0)) \\&\hspace{1.0cm} + \sum_{i = 1..T-1}\mathbb{E}_{(s_T,...,s_0) \sim \mu^{\pi_{\theta}}}[\log \frac{\mu^{\pi_{\theta}}(s_i)}{\mu^E(s_i)} ] \\
    & =  \mathbb{D}_{KL}(\mu^{\pi_{\theta}}(s_T,...,s_0)||\mu^E(s_T,...,s_0)) \\&\hspace{1.0cm} + \sum_{i = 1..T-1} \mathbb{D}_{KL}(\mu^{\pi_{\theta}}(s_i)||\mu^E(s_i))
\end{aligned}
\end{equation}

\subsection{Bounded rewards}\label{appendix:assumptions}
Since we use the SAC algorithm as a subroutine all rewards must be bounded.
This is true if all subterms of our reward function
\begin{equation}
\begin{aligned}
r(a_i, s_i) = & \mathbb{E}_{s_{i+1} \sim \mu^{\pi_{\theta}}(s_{i+1}|s_i)} [-\log p(s_{i+1}|a_i,s_i) \\& + \log \pi'_{\theta}(a_{i}|s_{i+1},s_i) + \log  \mu^{E}(s_{i+1}|s_i)]
\end{aligned}
\end{equation}
are bounded which holds if 
\begin{equation}
\begin{aligned}\epsilon \leq \pi'_{\theta}(a_{i}|s_{i+1},s_i), & p(s_{i+1}|a_i,s_i), \\& \mu^{E}(s_{i+1}|s_i) \leq H \hspace{0.5cm} \forall a_i,s_i,s_{i+1}
\end{aligned}
\end{equation}
for some $\epsilon$ and $H$ which is a rather strong assumption which requires compact action and state spaces and a non-zero probability to reach every state $s_{i+1}$ given any action $a_i$ from a predecessor state $s_i$.
Since this is in general not the case in practice we clip the logarithms of $\pi'_{\theta}(a_{i}|s_{i+1},s_i), p(s_{i+1}|a_i,s_i), \mu^{E}(s_{i+1}|s_i)$ to $[-15,1e9]$. It should be noted that clipping the logarithms to a maximum negative value still provides a reward signal which guides the imitation learning to policies which achieve higher rewards.

\subsection{Correctness of using replay buffer}\label{appendix:replaybuffer}
Here  we  argue that Algorithm \ref{alg:SOIL-TDM} leads to a local optimum of the KLD objective from Equation \ref{eq:kldsoiltdm} under the conditions: a) In the large sample limit per iteration b) appropriate density estimators and optimizers are used c) Equation \ref{eq:kldsoiltdm} is minimized by optimizing the policy using a maximum entropy RL algorithm d) the inverse action policy $\pi'(a|s',s)$ is trained using maximum likelihood from a replay buffer in an alternating fashion with the policy optimization. 

In Section \ref{sec:method} we show that Equation \ref{eq:kldsoiltdm} is a maximum entropy RL objective. Thus when optimizing the policy $\pi_{\theta(e)}$ in episode $e$ in Algorithm \ref{alg:SOIL-TDM} using the maximum entropy RL algorithm SAC \cite{haarnoja2018SAC} keeping the parameters of the conditional normalizing flows $\mu^E(s'|s)$, $\mu_{\phi}(s'|s, a)$, and $\mu_{\eta}(a|s', s)$ (to stay consistent with our method section, we use the distribution definitions here instead of the model definition) fixed which define the reward implies
\begin{equation}\label{eq:ineq1}
\begin{aligned}
\sum_{i = 0..T-1} & \mathbb{E}_{(s_i,a_i,s_{i+1}) \sim {\pi_{\theta(e)}}}[\log p(s_{i+1}|a_i,s_i) \\ & + \log \pi_{\theta(e)}(a_i|s_i) - \log \pi'_{\theta(e)}(a_{i}|s_{i+1},s_i) \\ & - \log  \mu^{E}(s_{i+1}|s_i)] \\
\leq \sum_{i = 0..T-1} & \mathbb{E}_{(s_i,a_i,s_{i+1}) \sim {\pi_{\theta(e-1)}}}[\log p(s_{i+1}|a_i,s_i) \\ & + \log \pi_{\theta(e-1)}(a_i|s_i) - \log \pi'_{\theta(e)}(a_{i}|s_{i+1},s_i) \\ & - \log  \mu^{E}(s_{i+1}|s_i)]
\end{aligned}
\end{equation}
Now, in the next episode $e+1$ the first part of Algorithm \ref{alg:SOIL-TDM}, i.e. optimizing the model of the inverse action policy $\pi'_{\theta(e+1)}$ with a maximum likelihood objective using the new replay buffer data $(s_i,a_i,s_{i+1}) \sim {\pi_{\theta(e)}}$ obtained from rollouts in episode $e+1$ with the new policy $\pi_{\theta(e)}$ trained in episode $e$  leads to
\begin{equation}\label{eq:ineq2}
\begin{aligned}
\sum_{i = 0..T-1} & \mathbb{E}_{(s_i,a_i,s_{i+1}) \sim {\pi_{\theta(e)}}}[ - \log \pi'_{\theta(e+1)}(a_{i}|s_{i+1},s_i) ] \\
\leq \sum_{i = 0..T-1} & \mathbb{E}_{(s_i,a_i,s_{i+1}) \sim {\pi_{\theta(e)}}}[ - \log \pi'_{\theta(e)}(a_{i}|s_{i+1},s_i) ]
\end{aligned}
\end{equation}
due to the maximum likelihood objective for the inverse action policy.

Using this inequality one obtains
\begin{equation}\label{eq:ineq3}
\begin{aligned}
\sum_{i = 0..T-1} & \mathbb{E}_{(s_i,a_i,s_{i+1}) \sim {\pi_{\theta(e)}}}[\log p(s_{i+1}|a_i,s_i) \\ & + \log \pi_{\theta(e)}(a_i|s_i) - \log \pi'_{\theta(e+1)}(a_{i}|s_{i+1},s_i) \\ & - \log  \mu^{E}(s_{i+1}|s_i)]
\\
\leq
\sum_{i = 0..T-1} & \mathbb{E}_{(s_i,a_i,s_{i+1}) \sim {\pi_{\theta(e-1)}}}[\log p(s_{i+1}|a_i,s_i) 
\\ & + \log \pi_{\theta(e-1)}(a_i|s_i) - \log \pi'_{\theta(e)}(a_{i}|s_{i+1},s_i)  
\\ & - \log  \mu^{E}(s_{i+1}|s_i)]
\end{aligned}
\end{equation}

Thus Algorithm \ref{alg:SOIL-TDM} optimizes Equation \ref{eq:kldsoiltdm} also in the "update dynamics models" part when training $\mu_{\eta}(a|s', s)$ using maximum likelihood from a replay buffer.
Thus, optimizing the policy $\pi$ using SAC and training the model of the inverse action policy $\pi'$ using the replay buffer and maximum likelihood are non-competing and non adversarial objectives, they alternately minimize the same objective in each part of Algorithm \ref{alg:SOIL-TDM} and decrease the Kulback-Leibler Divergence in each step, ending in a minimum at convergence since the KLD is bounded by $0$ from below.

The inequality from Equation \ref{eq:ineq2} is based on the maximum likelihood objective and a "clean" replay buffer that contains only samples from the current policy. But it can be shown that it also holds for a replay buffer which contains a mixture of samples from the current and old policies: the inequality holds for the mixture distribution which contains a fraction $\alpha$ of the replay buffer which stems from the new policy and a fraction $1-\alpha$ which stems from the old policies ($\alpha$ depends on the size of the replay buffer and the number of new samples obtained in the current rollout). I.e. $p(RB(e+1)) = \alpha \pi_{\theta (e+1)} + (1-\alpha)p(RB(e))$. Since the old inverse action policy $\pi'_{\theta(e)}$ is the argmax of the maximum likelihood objective of the $1-\alpha$ fraction of RB(e+1) which is RB(e) it is better or equal than any other inverse action policy with regard to that previous replay buffer RB(e). Thus
\begin{equation}\label{eq:ineq4}
\begin{aligned}
\\\sum_{i = 0..T-1} & \mathbb{E}_{(s_i,a_i,s_{i+1}) \sim {RB(e)}}[ - \log \pi'_{\theta(e+1)}(a_{i}|s_{i+1},s_i) ]
\\
\geq
\sum_{i = 0..T-1} & \mathbb{E}_{(s_i,a_i,s_{i+1}) \sim {RB(e)}}[ - \log \pi'_{\theta(e)}(a_{i}|s_{i+1},s_i) ]
\end{aligned}
\end{equation}
due to the maximum likelihood objective for $\pi'_{\theta(e+1)}$ on the data RB(e+1)) the following inequality follows:
\begin{equation}\label{eq:ineq5}
\begin{aligned}
\\\sum_{i = 0..T-1} & \mathbb{E}_{(s_i,a_i,s_{i+1}) \sim RB(e+1)}[ - \log \pi'_{\theta(e+1)}(a_{i}|s_{i+1},s_i) ]
\\
\leq
\sum_{i = 0..T-1} & \mathbb{E}_{(s_i,a_i,s_{i+1}) \sim RB(e+1)}[ - \log \pi'_{\theta(e)}(a_{i}|s_{i+1},s_i) ]
\end{aligned}
\end{equation}

Also, by using the mixture definition of RB(e+1) one can rewrite an Expectation over RB(e+1) as follows:
\begin{equation}
\begin{aligned}
\mathbb{E}_{(s_i,a_i,s_{i+1}) \sim RB(e+1)} [f] =  & \alpha \mathbb{E}_{(s_i,a_i,s_{i+1}) \sim \pi_{\theta(e)}} [f]
\\
& +(1-\alpha) \mathbb{E}_{(s_i,a_i,s_{i+1}) \sim RB(e)} [f]
\end{aligned}
\end{equation}
Using the expanded Expectation Inequality \ref{eq:ineq5} can be rewritten as follows:
\begin{equation}
\begin{aligned}
& \sum_{i = 0..T-1}  [ \alpha \mathbb{E}_{(s_i,a_i,s_{i+1}) \sim \pi_{\theta(e)}} - \log \pi'_{\theta(e+1)}(a_{i}|s_{i+1},s_i)
\\
& +(1-\alpha) \mathbb{E}_{(s_i,a_i,s_{i+1}) \sim RB(e)} - \log \pi'_{\theta(e+1)}(a_{i}|s_{i+1},s_i) ]
\\
\leq
& \sum_{i = 0..T-1}  [ \alpha \mathbb{E}_{(s_i,a_i,s_{i+1}) \sim \pi_{\theta(e)}} - \log \pi'_{\theta(e)}(a_{i}|s_{i+1},s_i)
\\
& +(1-\alpha) \mathbb{E}_{(s_i,a_i,s_{i+1}) \sim RB(e)} - \log \pi'_{\theta(e)}(a_{i}|s_{i+1},s_i) ]
\end{aligned}
\end{equation}
By using Inequality \ref{eq:ineq4} it can be rewritten to
\begin{equation}
\begin{aligned}
& \sum_{i = 0..T-1}  [ \alpha \mathbb{E}_{(s_i,a_i,s_{i+1}) \sim \pi_{\theta(e)}} - \log \pi'_{\theta(e+1)}(a_{i}|s_{i+1},s_i)
\\
& +(1-\alpha) \mathbb{E}_{(s_i,a_i,s_{i+1}) \sim RB(e)} - \log \pi'_{\theta(e+1)}(a_{i}|s_{i+1},s_i) ]
\\
\leq
& \sum_{i = 0..T-1}  [ \alpha \mathbb{E}_{(s_i,a_i,s_{i+1}) \sim \pi_{\theta(e)}} - \log \pi'_{\theta(e)}(a_{i}|s_{i+1},s_i)
\\
& +(1-\alpha) \mathbb{E}_{(s_i,a_i,s_{i+1}) \sim RB(e)} - \log \pi'_{\theta(e+1)}(a_{i}|s_{i+1},s_i) ]
\end{aligned}
\end{equation}
which implies (by subtracting the common $1-\alpha$ term from both sides) 

\begin{equation}
\begin{aligned}
& \sum_{i = 0..T-1}  [ \alpha \mathbb{E}_{(s_i,a_i,s_{i+1}) \sim \pi_{\theta(e)}} - \log \pi'_{\theta(e+1)}(a_{i}|s_{i+1},s_i)
\\
\leq
& \sum_{i = 0..T-1}  [ \alpha \mathbb{E}_{(s_i,a_i,s_{i+1}) \sim \pi_{\theta(e)}} - \log \pi'_{\theta(e)}(a_{i}|s_{i+1},s_i)
\end{aligned}
\end{equation}

which is Inequality \ref{eq:ineq2} multiplied by $\alpha$ and thus also implies (together with Inequality \ref{eq:ineq1}) Inequality \ref{eq:ineq3}. From this follows the convergence of Algorithm \ref{alg:SOIL-TDM} to a minimum when using a mixed replay buffer.

\subsection{Implementation Details}\label{sec:implDet}
We use the same policy implementation for all our SOIL-TDM experiments. The stochastic policies $\pi_{\theta}(a_t|s_t)$ are modeled as a diagonal Gaussian to estimate continuous actions with two hidden layers (512, 512) with ReLU nonlinearities. 

To train a policy using SAC as the RL algorithm, we also need to model a Q-function. Our implementation of SAC is based on the original implementation from Haarnoja et al. \shortcite{haarnoja2018sacapps} and the used hyperpameter are described in Table \ref{tb:sachyp}. In this implementation, they use two identical Q-Functions with different initialization to stabilize the training process. These Q-Functions are also modeled with an MLP having two hidden layers (512, 512) and Leaky ReLU. We kept the entropy parameter $\alpha$ fixed to 1 and did not apply automatic entropy tuning as described by Haarnoja et al. \shortcite{haarnoja2018sacapps}. 

\begin{table}[ht]
	\renewcommand{\arraystretch}{1.3}
	\centering
	\caption{Training Hyperparameter}\label{tb:sachyp}
	\begin{small}
	\begin{tabular}{l|l}
	    \hline
		SAC Parameter & Value \\\hline
		Optimizer & Adam\nocite{KingmaB14} \\
		learning rate policy & $1 \cdot 10^{-4}$ \\
		learning rate Q-function & $3 \cdot 10^{-4}$ \\
		discount $\gamma$ (Halfcheetah, Walker2D) & $0.7$ \\
		discount $\gamma$ & $0.9$ \\
		mini batch size & $2048$ \\
		replay buffer size & $1 \cdot 10^5$\\
		target update interval & $1$ \\
		number of environments & $16$ \\
		max number of environment steps & $4.0 \cdot 10^6$ \\
		\\\hline
		SOIL-TDM Parameter & Value \\\hline
		expert transition model training steps & $10^{3} - 10^{4}$ \\
		learning rate expert transition model & $1 \cdot 10^{-4}$ \\
		learning rate forward dynamics model & $1 \cdot 10^{-4}$ \\
		learning rate backward dynamics model & $1 \cdot 10^{-4}$ \\
		update interval dynamics models & $1$ \\
	\end{tabular}
	\end{small}
\end{table}

We implement all our models for SOIL-TDM using the PyTorch framework version 1.9.0\footnote{https://github.com/pytorch/pytorch}.  
To estimate the imitation reward in SOIL-TDM a model for the expert transitions $\mu_E(s'|s)$ as well as a forward $\mu_{\phi}(s'|s, a)$ and backward dynamics model $\mu_{\eta}(a|s', s)$ has to be learned. All three density models are based on RealNVPs \cite{realNVP} consisting of several flow blocks where MLPs preprocess the conditions to a smaller condition size. The RealNVP transformation parameters are also calculated using MLPs, which process a concatenation of the input and the condition features. After each flow block, we added activation normalization layers like \cite{NEURIPS2018_d139db6a}. To implement these models, we use the publicly available VLL-FrEIA\footnote{https://github.com/VLL-HD/FrEIA (Open Source MIT License)} framework version 0.2 \cite{ardizzone2018analyzing} using their implementation of GLOWCouplingBlocks with exponent clamping activated. The setup for each model is layed out in Table \ref{tb:flowSetup}. We add Gaussian noise to the state vector as a regularizer for the training of the expert transition model $\mu_E(s'|s)$ which increased training stability for low amount of expert trajectories. We implement a linear decrease of the standard deviation from 0.05 to 0.005 during the training of the expert model. 

We train and test all algorithms on a computer with 8 CPU cores, 64 GB of working memory and an RTX2080 Ti Graphics card. The compute time for the SOIL-TDM method depends on the time to convergence and is from 4h to 14h.  

\begin{table}[ht]
	\renewcommand{\arraystretch}{1.3}
	\centering
	\caption{Normalizing Flow Setup}\label{tb:flowSetup}
	\begin{small}
	\begin{tabular}{l|c|c|c}
		 & $\mu^E(s'|s)$ & $\mu_{\phi}(s'|s, a)$ & $\mu_{\eta}(a|s', s)$ \\\hline
		N flow blocks & 16 & 16 & 16  \\
		Condition hidden neurons & 64 & 48 & 256 \\
		Condition hidden layer & 2 & 2 & 2 \\
		Condition feature size & 32 & 32 & 32 \\
		Flow block hidden neurons & 64 & 48 & 256 \\
		Flow block hidden layer & 2 & 2 & 2 \\
		Exponent clamping & 6 & 1 & 1 \\
	\end{tabular}
	\end{small}
\end{table}

\subsection{Expert Data Generation and Baseline Methods}\label{appendix:expSetup}
The expert data is generated by training an expert policy based on conditional normalizing flows and the SAC algorithm on the environment reward. A conditional normalizing flow policy has been chosen for the expert to make the distribution matching problem for the baseline methods and SOIL-TDM - which employ a conditional Gaussian policy - more challenging and more similar to real-world IL settings. The idea is that real-world demonstrations might me more complex and experiments using the same policy setup for the expert and the imitator might not well translate to real-world tasks. 

The stochastic flow policy $\pi_{\theta}(a_t|s_t)$ is based on RealNVPs \cite{realNVP} which have the same setup as the normalizing flow implementations used for $\mu^E(s'|s)$, $\mu_{\phi}(s'|s, a)$, and $\mu_{\eta}(a|s', s)$ also using N=16 GLOWCouplingBlocks. The state is processed as a condition with one MLP having a hidden size of 128. Each flow block has an additional fully connected layer to further process the condition to a small feature size of 8. Every flow block has 128 hidden neurons. Finally, each action is passed through a tanh layer as it is described in the SAC implementation. The log probability calculation was adapted accordingly \cite{haarnoja2018sacapps}. The final episode reward of the trained expert policy is in Table \ref{tb:expertR}. 

The expert trajectories are generated using the trained policy and saved as done by \cite{GAIL}\footnote{ https://github.com/openai/imitation (Open Source MIT License)}. For the OPOLO\footnote{https://github.com/illidanlab/opolo-code (Open Source MIT License)} and F-IRL\footnote{https://github.com/twni2016/f-IRL (Open Source MIT License)} baseline, the original implementations with the official default parameters for each environment are used. Only the loading of the expert data was changed to use the demonstrations of the previously trained normalizing flow policy. Since, no official code for FORM was publicly available, the FORM baseline was implemented based on our method. We changed the reward to use the state prediction effect model $\mu^{\pi_{\theta}}_{\Phi}(s'|s)$ as proposed by Jaegle et al. \shortcite{jaegle2021imitation}. Both effect models where implemented using the same implementation as for our expert transition model $\mu^E(s'|s)$. Training of the expert effect model $\mu^{E}_{\omega}(s'|s)$ was performed offline with the same hyperparameter setup used for our expert transition model $\mu^E(s'|s)$ (see Table \ref{tb:sachyp} and Table \ref{tb:flowSetup}). The policy optimization was done using the same SAC setup as for our method since FORM does not depend on a specific RL algorithm \cite{jaegle2021imitation}. We tested different setups for the entropy parameter $\alpha$ and found that automatic entropy tuning as described by Haarnoja et al. \shortcite{haarnoja2018sacapps} worked best. 

\begin{table}[ht]
	\renewcommand{\arraystretch}{1.3}
	\centering
	\caption{Episode Reward of Expert Policy}\label{tb:expertR}
	\begin{small}
	\begin{tabular}{l|c}
		Environment & Average Expert Episode Reward \\\hline
		AntBulletEnv-v0 & 2583 \\
		HalfCheetahBulletEnv-v0 & 2272  \\
		HopperBulletEnv-v0 & 2357 \\
		Walker2DBulletEnv-v0 & 1805 \\
		HumanoidBulletEnv-v0 & 2750 \\
	\end{tabular}
	\end{small}
\end{table}

For our experiment with unkown true environment reward
the following selection criteria are used. For "OPOLO est. reward" the estimated return based on the reward $r(s,s')=-log(1-D(s,s'))$ with the state $s$, next state $s'$ and the discriminator output $D(s,s')$ is used. For "OPOLO pol. loss" the original OPOLO policy loss is used: 
\begin{equation}
\begin{aligned}
   J(\pi_{\theta}, Q) =  (1-\gamma) & \mathbb{E}_{s\sim S_0}[Q(s,\pi_{\theta}(s))] + \\
   & \mathbb{E}_{(s,a,s')\sim R}[f(r(s,s')+ \\
   & \gamma Q(s',\pi_{\theta}(s'))-Q(s,a))]
\end{aligned}
\end{equation}
With the Q-Function $Q$ and the $f$-divergence function. For both estimates the original OPOLO implementation was used. For FORM the convergence was estimated with the estimated return based on the reward: $ r_t = log \mu^E_{\omega}(s'|s) - log \mu^{\pi_{\theta}}_{\Phi}(s'|s) $ using the normalizing flow effect models. For F-IRL we use the implementation of Ni et al. \shortcite{firl2020corl} for the estimate of the Jensen-Shannon divergence between state marginals of the expert and policy
\begin{equation*}
\frac{1}{2}\int p(x)log\frac{2p(x)}{p(x)+q(x)}+q(x)log\frac{2q(x)}{p(x)+q(x)}dx.
\end{equation*}

\subsection{Ablation Study}\label{sec:ablStud}

In this section we want to investigate the influence of difference components for our proposed imitation learning setup. First we want to answer the question if learning additional backward and forward dynamics models to estimate the state transition KLD improves policy performance. We compare our proposed method SOIL-TDM to an approach where we train a policy only based on the log-likelihood of the expert defined by:
\begin{equation} \label{eq:ablReward}
r_{abl}(s,s')=\log \mu^E(s'|s)    
\end{equation}

\begin{figure}
\includegraphics[width=0.49\textwidth]{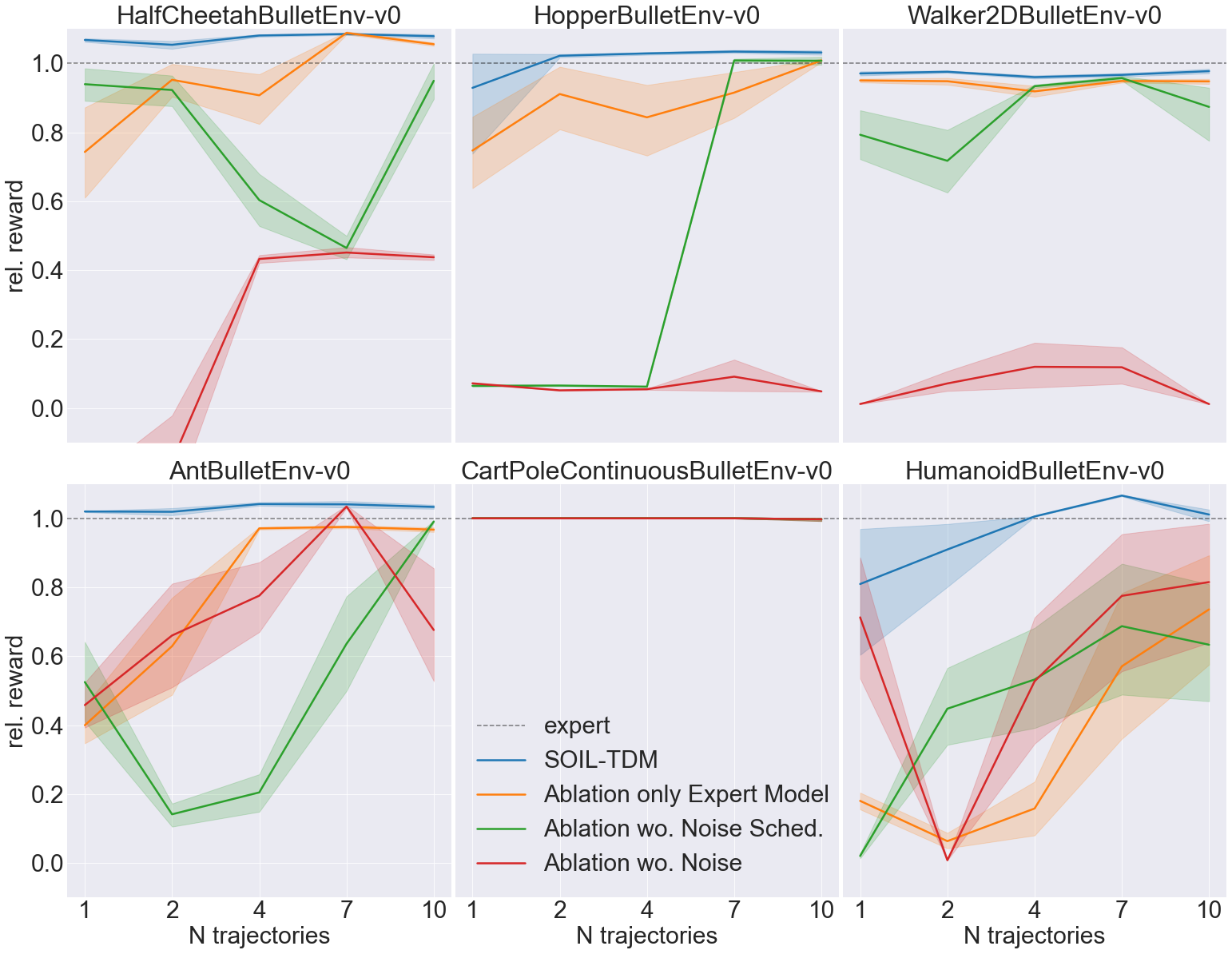}
\caption{Best environment reward for ablation experiments. Relative reward for different amount of expert trajectories. The value 1 corresponds to expert policy performance.}
\label{fig:reward_abl_plot}
\end{figure}

Using this reward the policy is optimized using SAC as described earlier. We call this simplified reward design approach "Ablation only Expert Model". By comparing the performance of this method to our approach, we can show that learning additional density models to estimate forward and backward dynamics leads to improved policy performance. The resulting rewards are plotted in Figure \ref{fig:reward_abl_plot}. The relative reward using this ablation method is much lower compared to SOIL-TDM. Only for a high amount of trajectories does this method reach expert-level performance.

\begin{table}
	\renewcommand{\arraystretch}{1.3}
	\centering
	\caption{Test log-likelihood values of expert transition models $\mu^E(s'|s)$ for 1, 2, 4, 7, and 10 training trajectories using 20 unknown test trajectories.}\label{tb:abl1}
	\begin{small}
	\begin{tabular}{l|c}
		Environment & Log-Likelihood for $\mu^E(s'|s)$  \\\hline
		Ant & \makecell{$38.5, 43.5, 54.2, 59.9, 62.7$} \\
		HalfCheetah & \makecell{$45.5,47.8,70.2, 71.1,71.3$} \\
		Hopper & \makecell{$14.4,13.8,15.9, 40.7,36.9$}  \\
		Walker & \makecell{$38.9,47.4,62.5, 65.2,67.4$} \\
		Humanoid &  \makecell{$39.5,50.9,62.9,77.7,77.4$}\\
	\end{tabular}
	\end{small}
\end{table}

We furthermore want to evaluate how the quality of the learned normalizing flows affects the overall algorithm performance. We therefore report the estimated test log-likelihood of the trained expert models $\mu^E(s'|s)$ for different amount of expert trajectories using a separate test dataset with 20 unseen expert trajectories in table \ref{tb:abl1}. The influence of the expert model quality on the overall algorithm performance can be seen by comparing the test log-likelihood to the reward of the trained policy (in Figure \ref{fig:reward_abl_plot} "SOIL-TDM" for our method and "Ablation only Expert Model" for the simplified reward using the same expert model $\mu^E(s'|s)$). SOIL-TDM is less sensitive to expert model performance compared to "Ablation only Expert Model".  

\begin{table}
	\renewcommand{\arraystretch}{1.3}
	\centering
	\caption{Test log-likelihood values of expert transition models $\mu_{abl1}^E(s'|s)$ for 1, 2, 4, 7, and 10 training trajectories using 20 unknown test trajectories.}\label{tb:abl2}
	\begin{small}
	\begin{tabular}{l|c}
		Environment & Log-Likelihood for $\mu_{abl1}^E(s'|s)$ (constant noise) \\\hline
		Ant & \makecell{$30.9,30.8,41.1, 41.1,59.0$} \\
		HalfCheetah &  \makecell{$26.3,25.1,78.3, 77.9,79.2$}  \\
		Hopper & \makecell{$-31.6,-26.9,-24.1 ,41.6,40.8$} \\
		Walker & \makecell{$42.1,46.1,55.8,56.8,58.1$} \\
		Humanoid & \makecell{$26.4,38.0,51.5,54.6,57.4$} \\
	\end{tabular}
	\end{small}
\end{table}

\begin{table}
	\renewcommand{\arraystretch}{1.3}
	\centering
	\caption{Test log-likelihood values of expert transition models $\mu_{abl2}^E(s'|s)$ for 1, 2, 4, 7, and 10 training trajectories using 20 unknown test trajectories.}\label{tb:abl3}
	\begin{small}
	\begin{tabular}{l|c}
		Environment & Log-Likelihood for $\mu_{abl2}^E(s'|s)$ (no noise) \\\hline
		Ant & \makecell{$28.7,31.9,37.3, 37.4,38.9$} \\
		HalfCheetah & \makecell{$-inf,-inf,-270.2, -194.2,-187.8$} \\
		Hopper & \makecell{$-171.3,-629.0,-379.1, -87.1,-70.9$} \\
		Walker & \makecell{$-inf,-inf,-389.0, -182.4,-180.1$} \\
		Humanoid & \makecell{$25.9,36.7,46.4,50.1,52.1$}\\
	\end{tabular}
	\end{small}
\end{table}

Lastly, we want to investigate the expert model training. In the improved expert model training Routine, we regularize the optimization by adding Gaussian noise to the expert state values and linearly decrease its standard deviation down to $0.005$ during training. As an additional ablation, we tested 2 different training setups for the expert transition models $\mu^E(s'|s)$. For the first model ($\mu_{abl1}^E(s'|s)$) we omit the noise decay and only use the final constant Gaussian noise during training. For the second model ($\mu_{abl2}^E(s'|s)$) no noise is added during training of the expert model. We evaluated also the test log-likelihood of the trained models using the test dataset with 20 unseen expert trajectories. The resulting test log-likelihoods are in Tables \ref{tb:abl2} - \ref{tb:abl3}. The results show that, constant Gaussian noise already improves the performance of the expert model and our applied noise scheduling routine results in further performance improvements. We also used the trained expert models for policy training based on the ablation reward from Equation \ref{eq:ablReward}. These ablation methods are called "Ablation wo. Noise Sched.", "Ablation wo. Noise" respectively. The final rewards of these methods are also plotted in Figure \ref{fig:reward_abl_plot}. The results indicate that adding noise to the states during the offline training of the expert transition model also improves final policy performance.

\subsection{Additional Results}\label{sec:addRes}

The following figures (Figure \ref{fig:tf1} - \ref{fig:tf2}) show the policy loss and the estimated reward together with the environment reward during the training on different pybullet environments for OPOLO, F-IRL, FORM, and SOIL-TDM (our method). All plots have been generated from training runs with 4 expert trajectories and 10 test rollouts. It can be seen that the estimated reward and policy loss from SOIL-TDM correlates well with the true environment reward. It is possible that the policy loss of SOIL-TDM is lower than 0 since its based not on the true distributions. Instead its based on learned and inferred estimates of expert state conditional distribution, policy state conditional distribution, policy inverse action distribution and q-function with relatively large absolute values ($\sim50-100$) each. These estimation errors accumulate in each time-step due to sum and subtraction and due to the Q-function also over (on average) 500 timesteps which can lead to relatively large negative values.

\begin{figure*}[t]
\begin{center}
\includegraphics[width=0.8\textwidth]{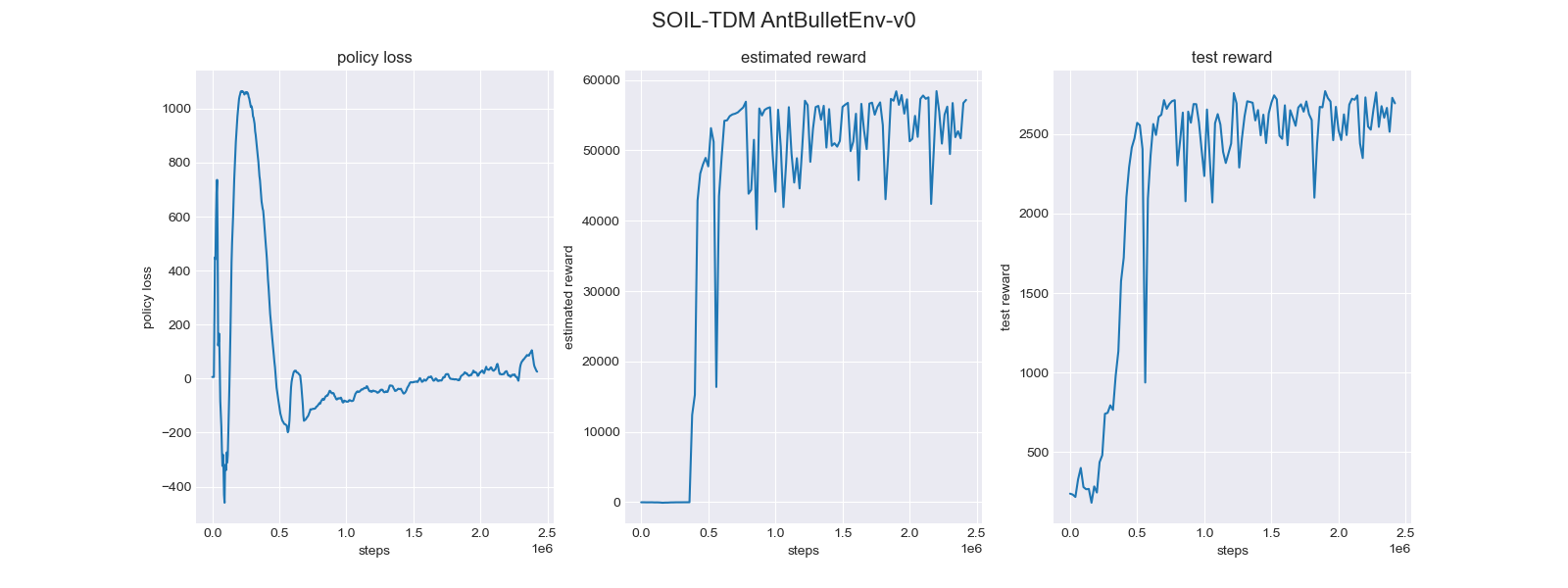}
\includegraphics[width=0.8\textwidth]{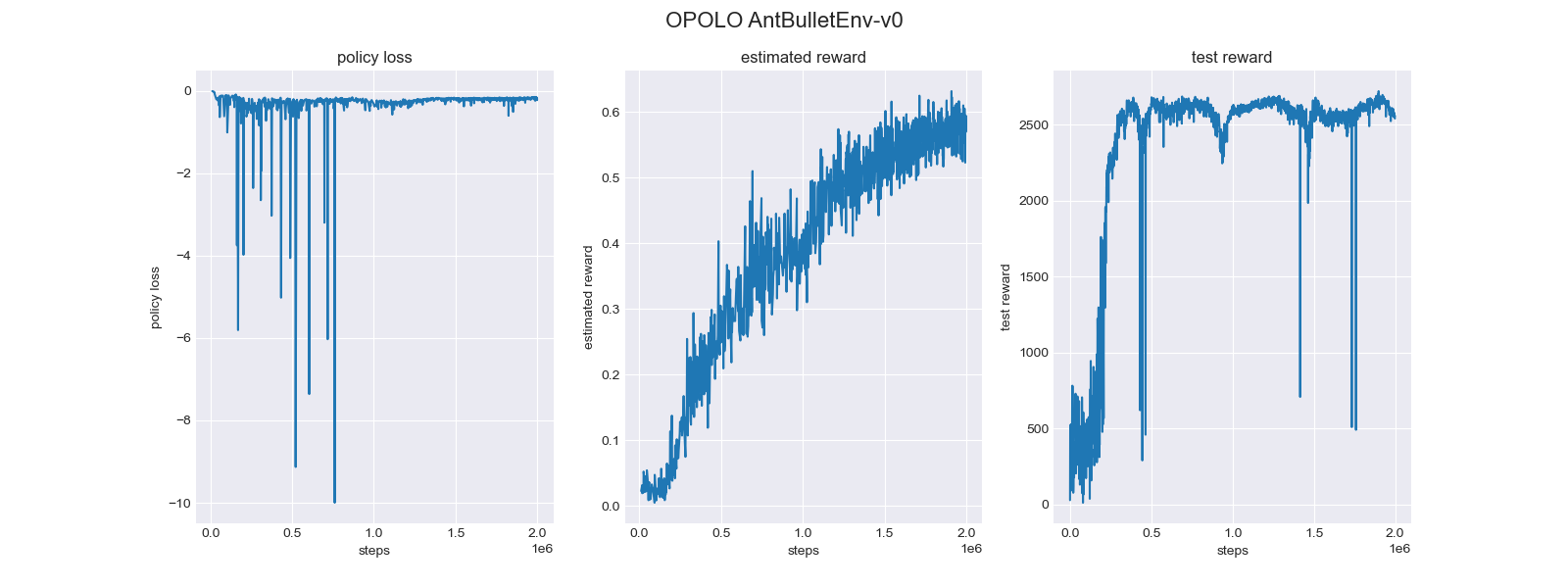}
\includegraphics[width=0.8\textwidth]{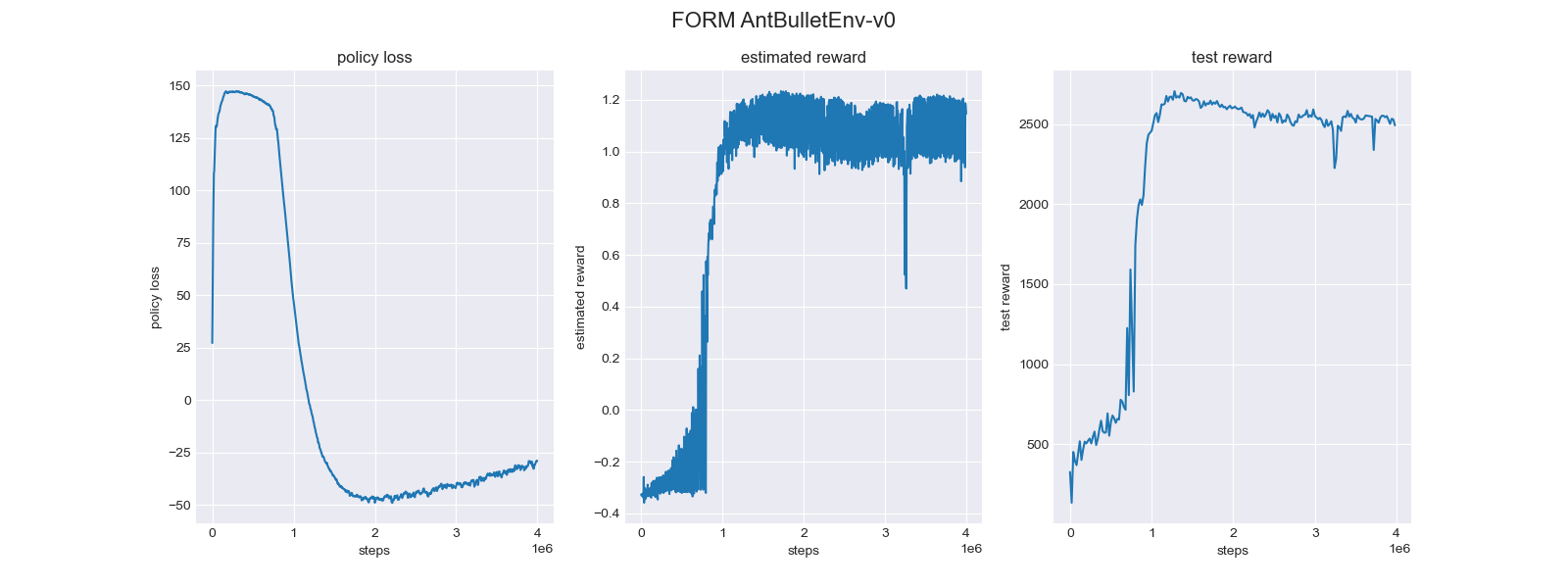}
\includegraphics[width=0.8\textwidth]{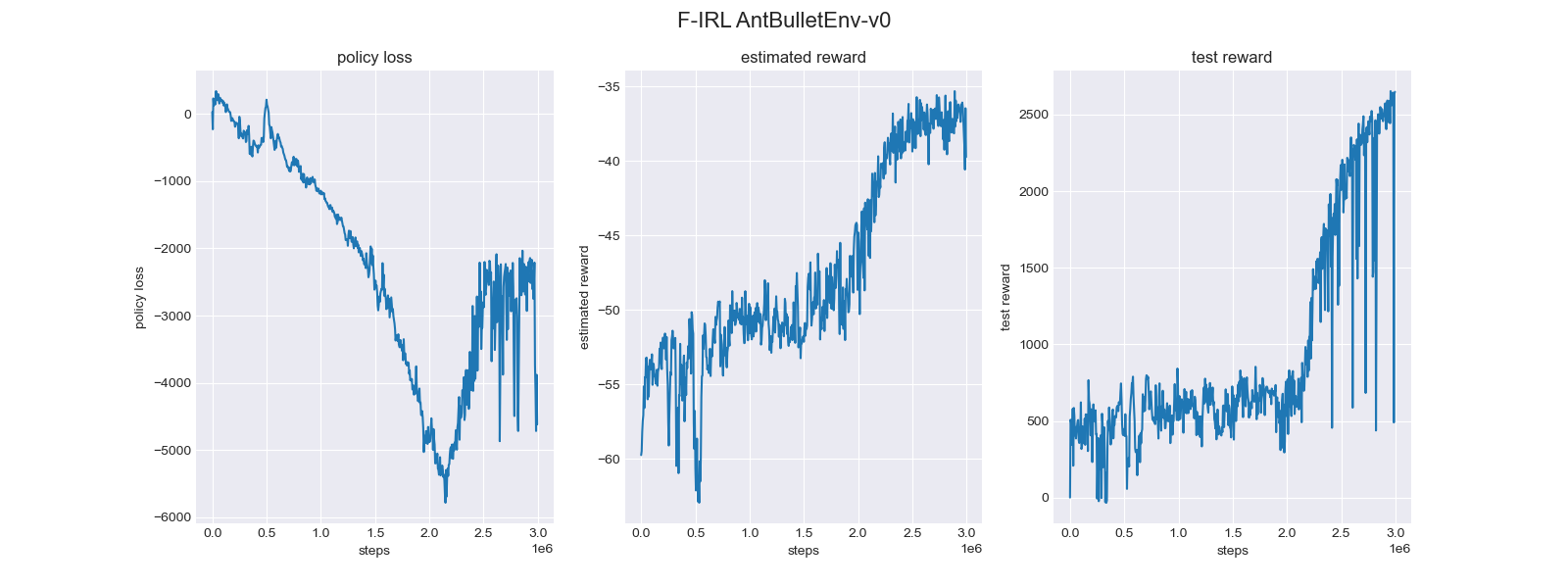}
\caption{The policy loss, estimated reward and the environment test loss during training in the pybullet Ant environment using our proposed SOIL-TDM and the OPOLO, F-IRL, and FORM implementations with 4 expert trajectories.}
\label{fig:tf1}
\end{center}
\end{figure*}

\begin{figure*}[t]
\begin{center}
\includegraphics[width=0.8\textwidth]{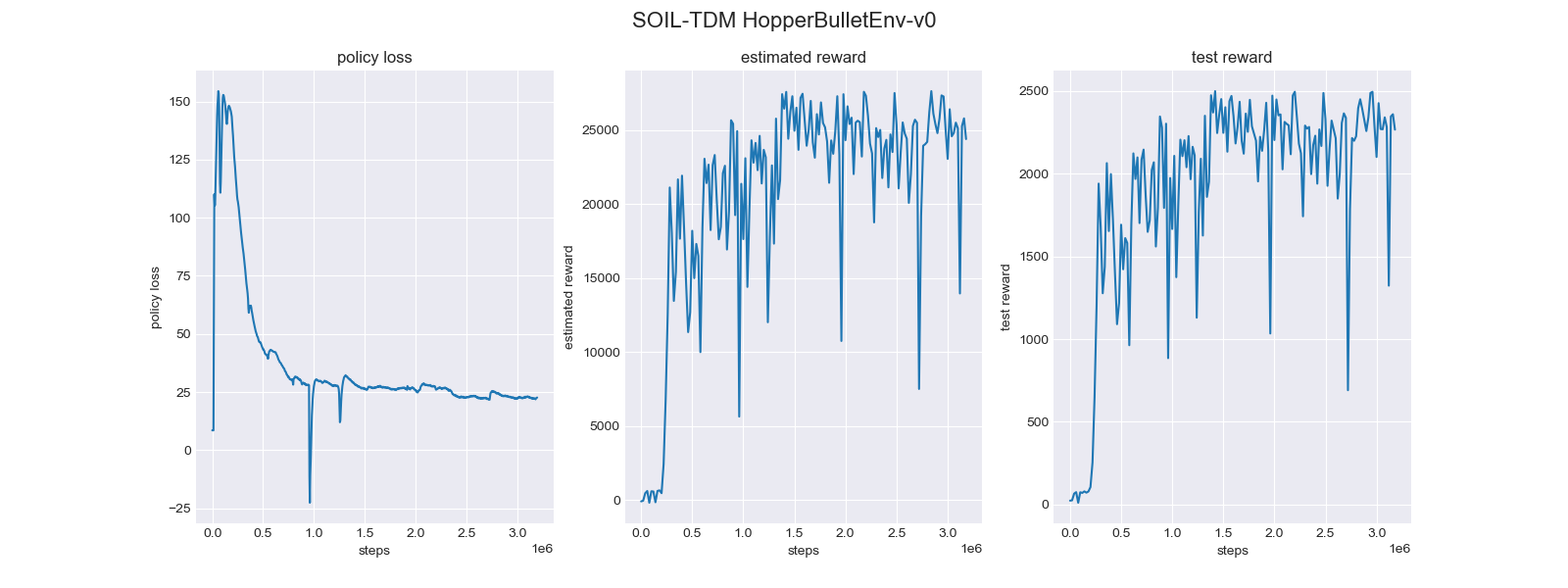}
\includegraphics[width=0.8\textwidth]{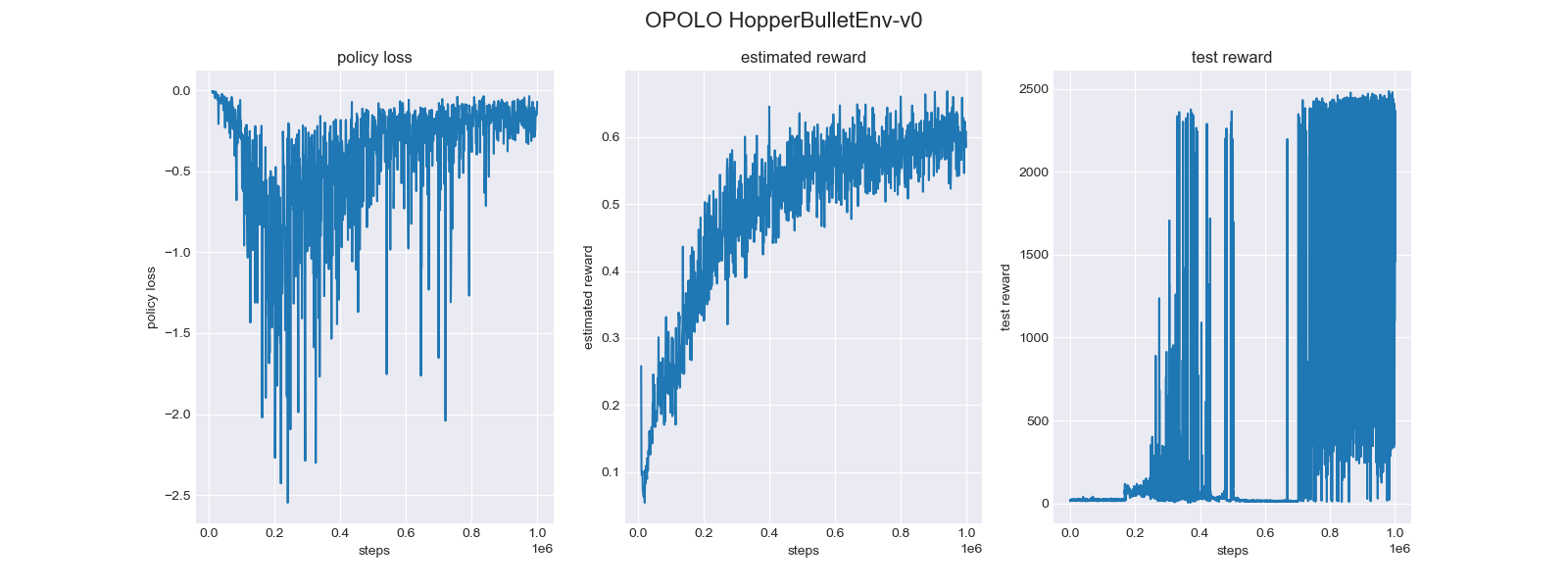}
\includegraphics[width=0.8\textwidth]{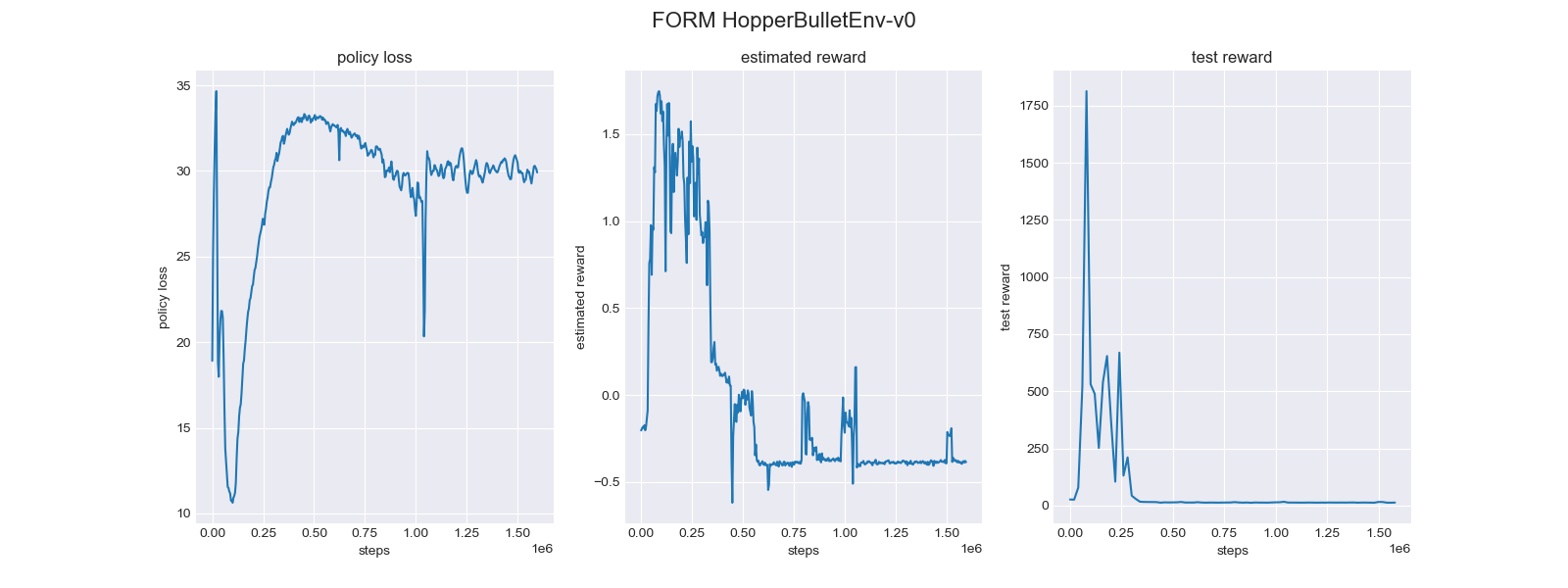}
\includegraphics[width=0.8\textwidth]{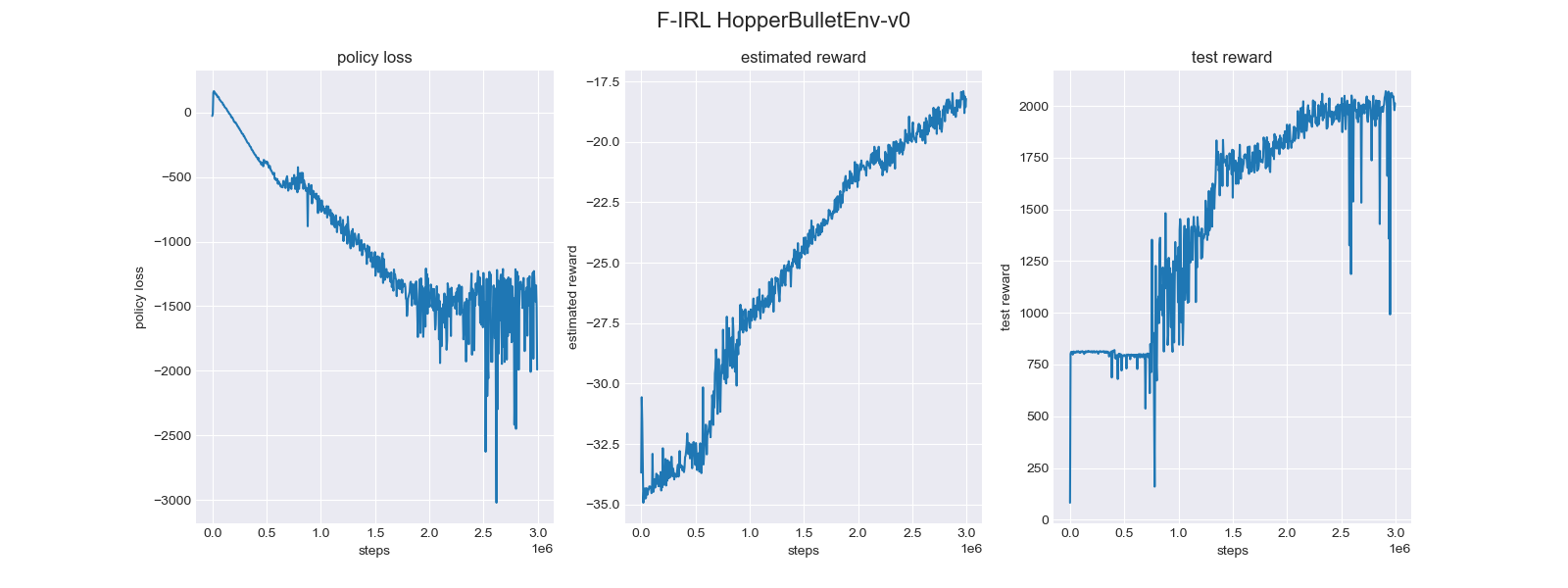}
\caption{The policy loss, estimated reward and the environment test reward during training in the pybullet Hopper environment using our proposed SOIL-TDM and the OPOLO, F-IRL, and FORM implementations with 4 expert trajectories.}
\label{fig:tf7}
\end{center}
\end{figure*}

\begin{figure*}[t]
\begin{center}
\includegraphics[width=0.8\textwidth]{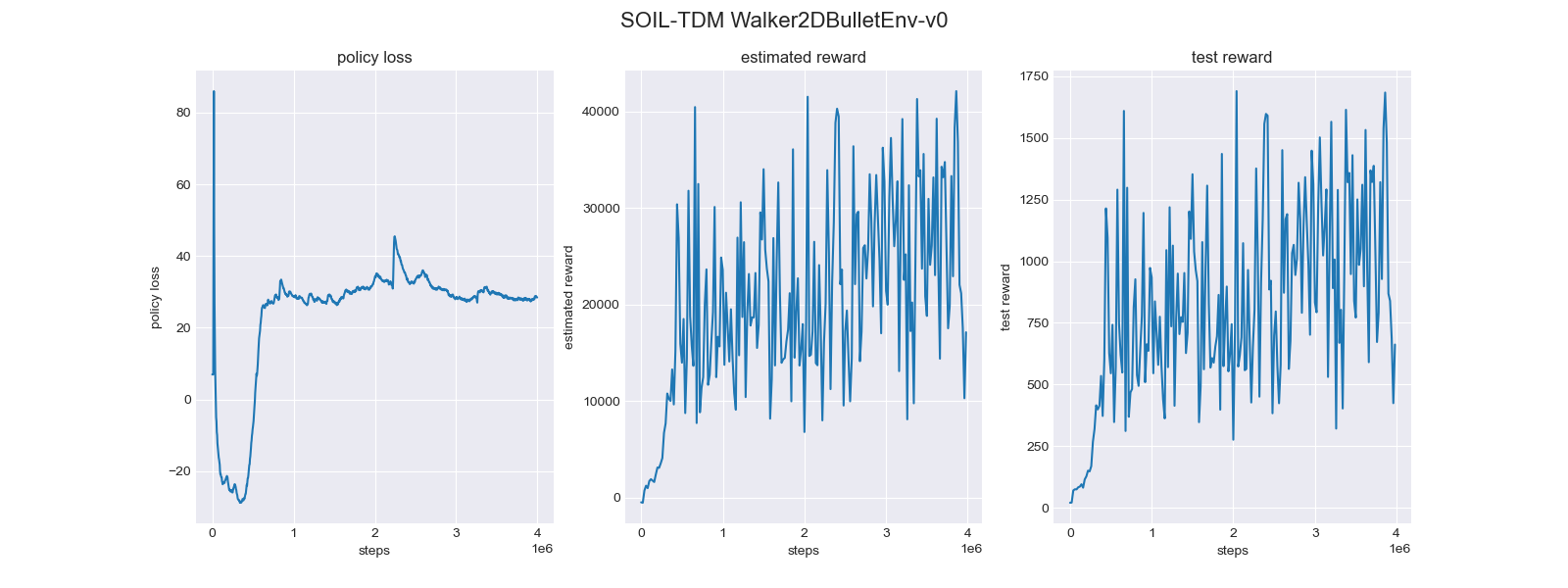}
\includegraphics[width=0.8\textwidth]{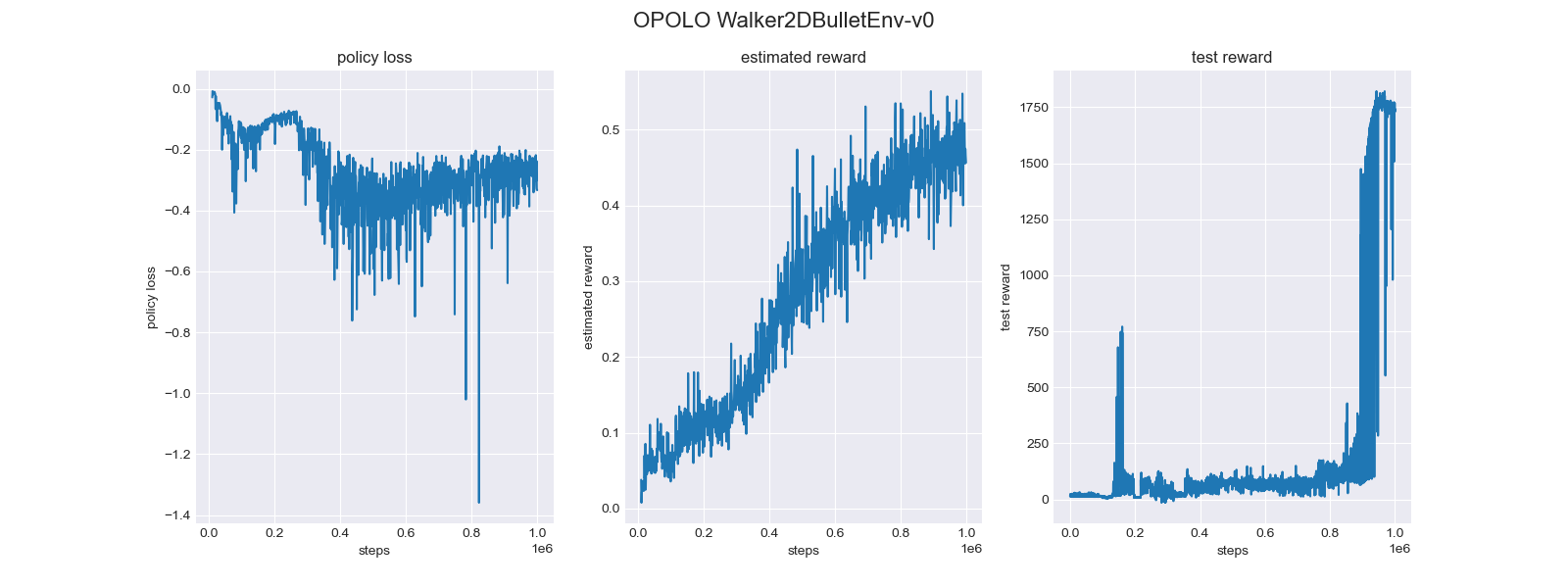}
\includegraphics[width=0.8\textwidth]{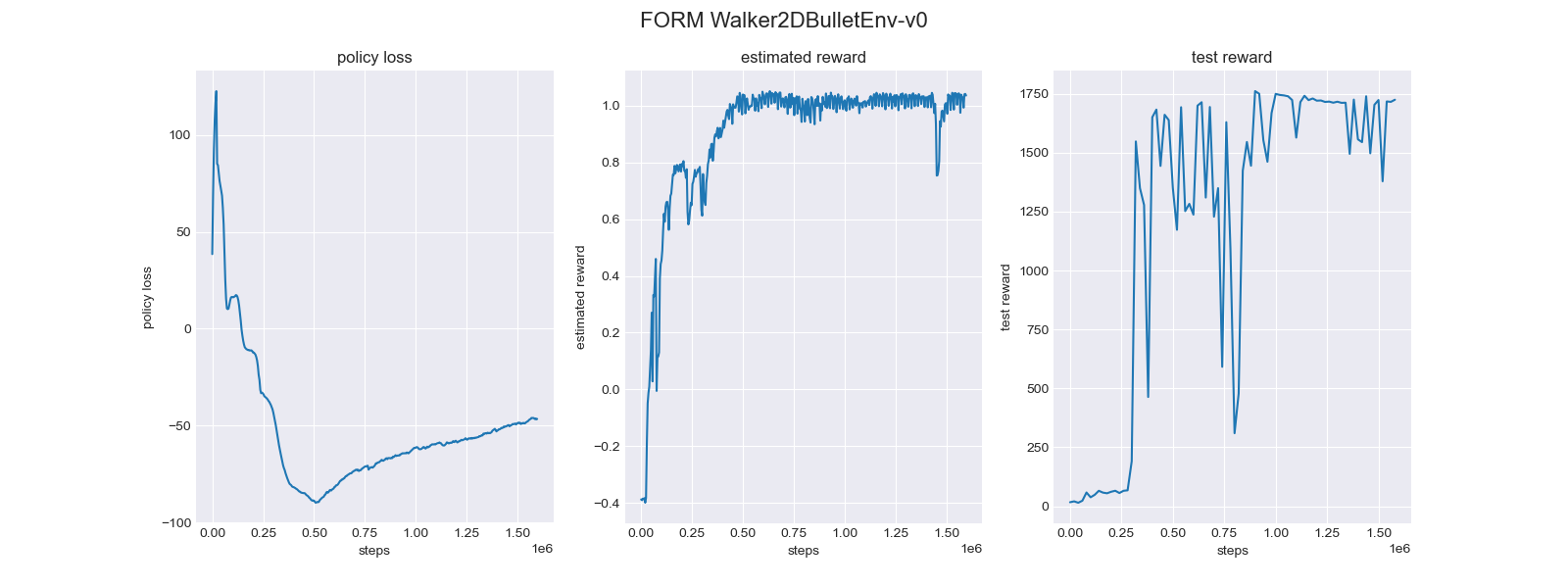}
\includegraphics[width=0.8\textwidth]{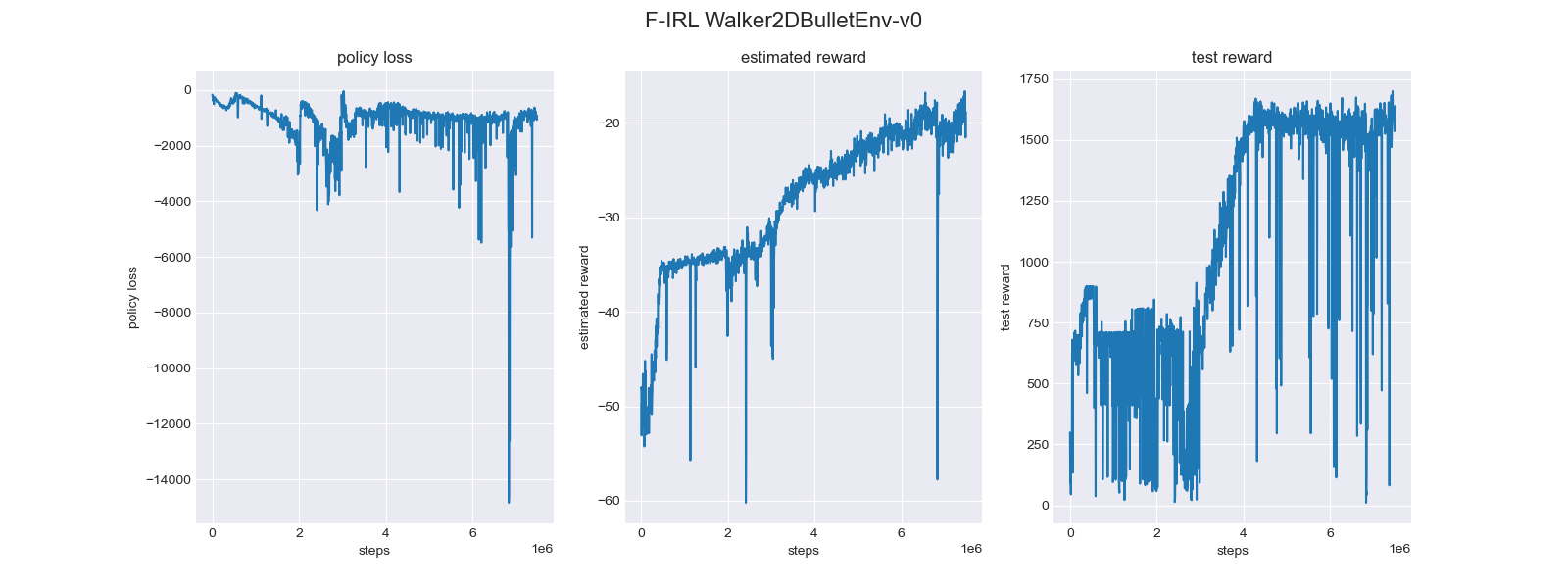}
\caption{The policy loss, estimated reward and the environment test reward during training in the pybullet Walker2D environment using our proposed SOIL-TDM and the OPOLO, F-IRL, and FORM implementations with 4 expert trajectories.}
\label{fig:tf3}
\end{center}
\end{figure*}

\begin{figure*}[t]
\begin{center}
\includegraphics[width=0.8\textwidth]{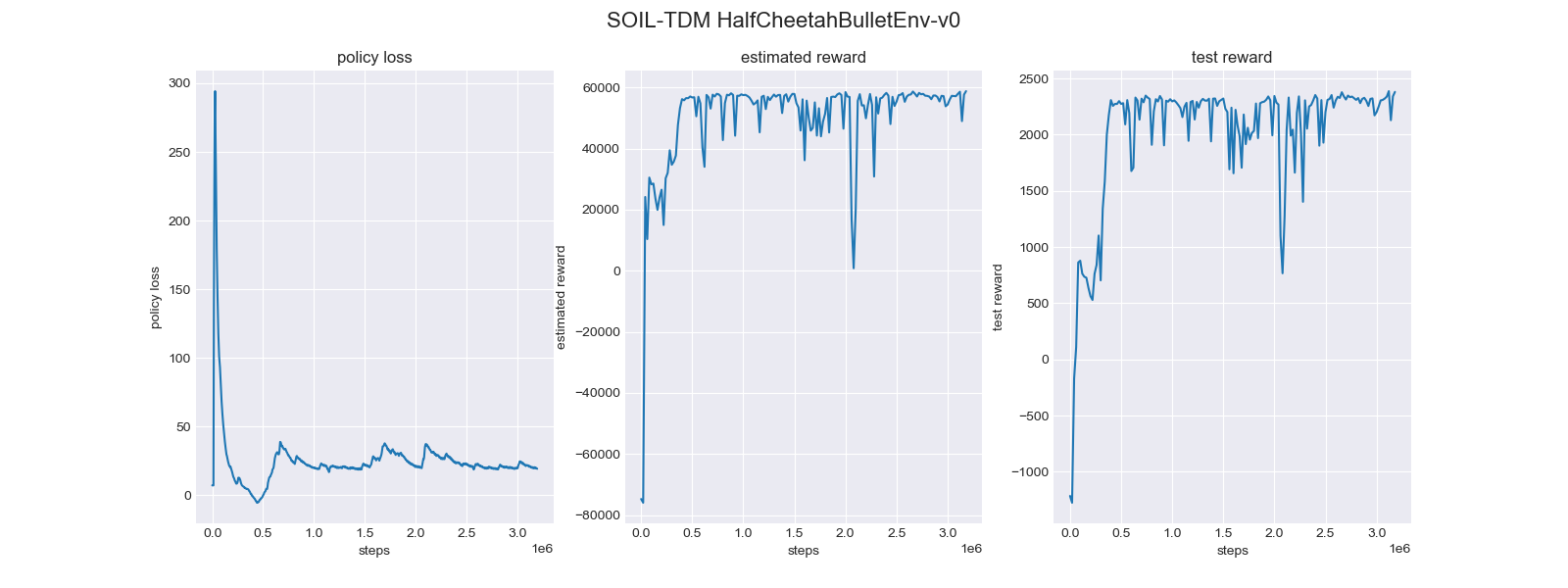}
\includegraphics[width=0.8\textwidth]{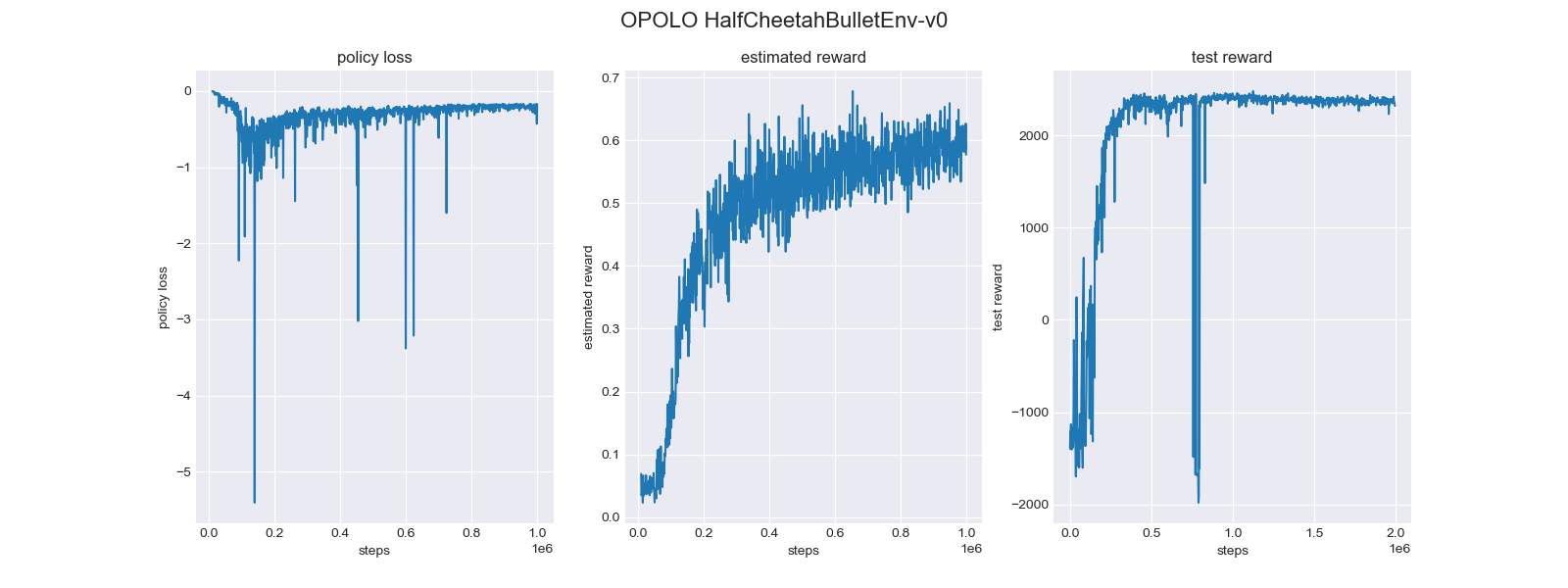}
\includegraphics[width=0.8\textwidth]{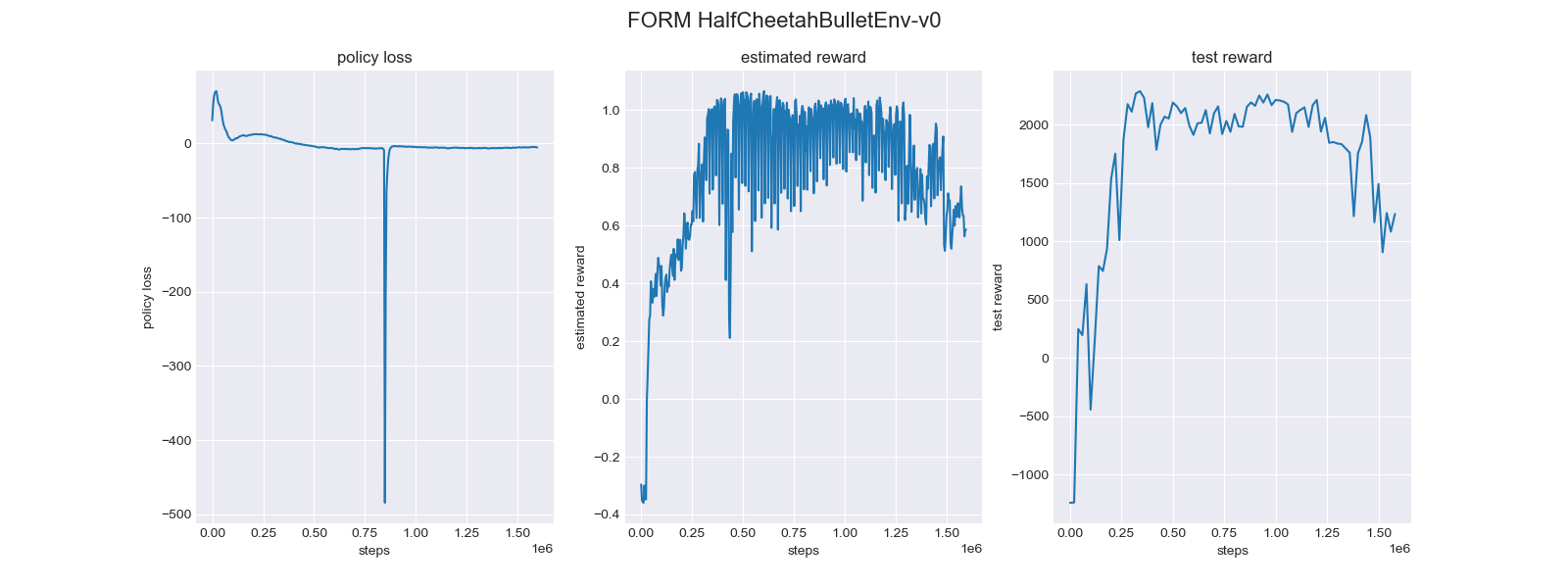}
\includegraphics[width=0.8\textwidth]{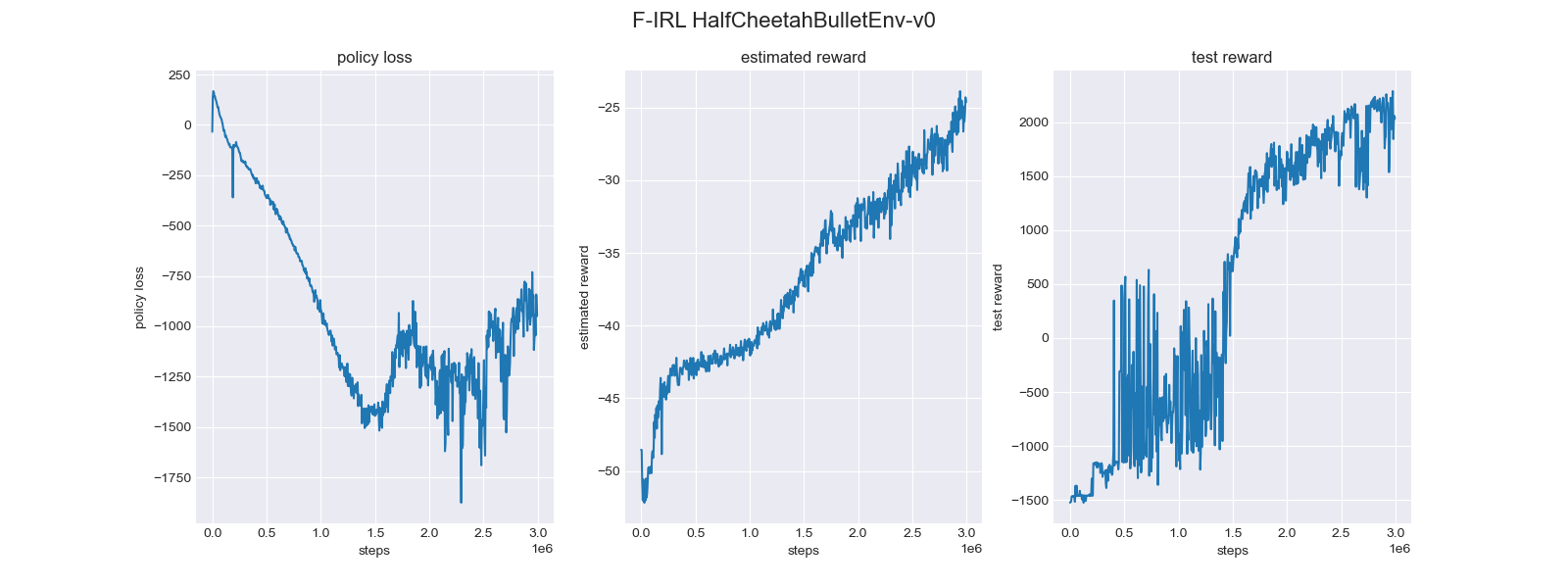}
\caption{The policy loss, estimated reward and the environment test reward during training in the pybullet HalfCheetah environment using using our proposed SOIL-TDM and the OPOLO, F-IRL, and FORM implementations with 4 expert trajectories.}
\label{fig:tf4}
\end{center}
\end{figure*}

\begin{figure*}[t]
\begin{center}
\includegraphics[width=0.8\textwidth]{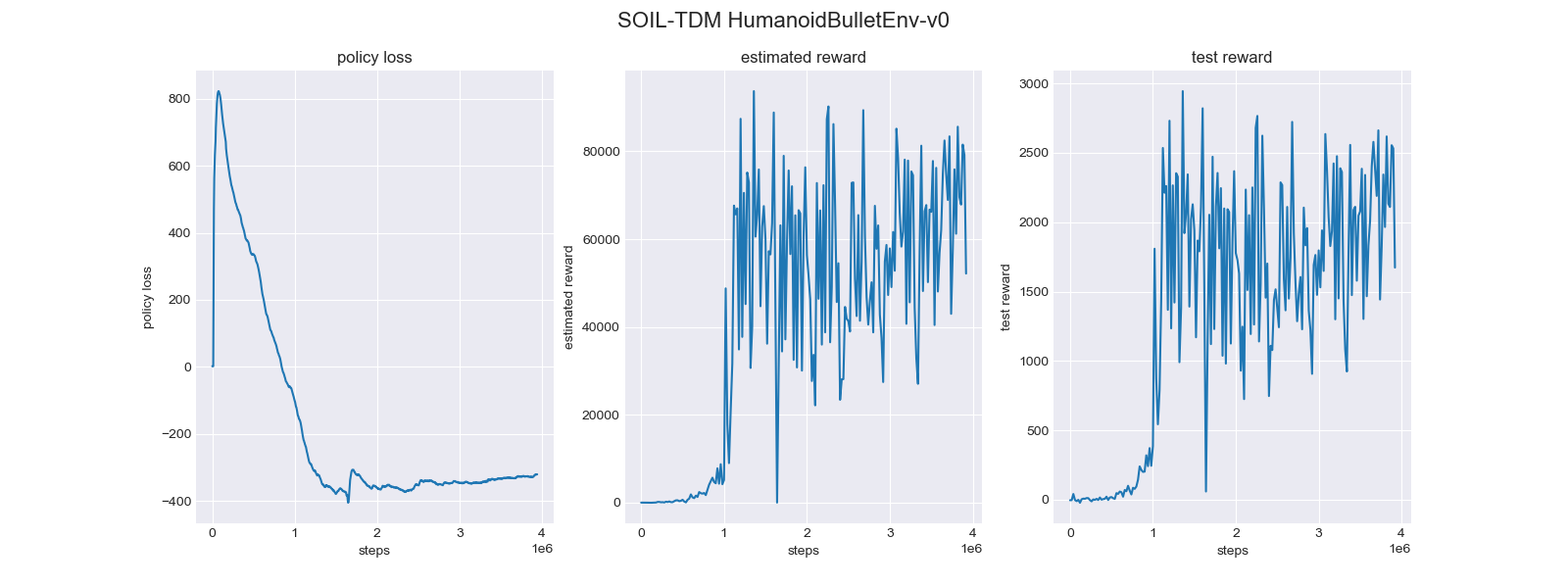}
\includegraphics[width=0.8\textwidth]{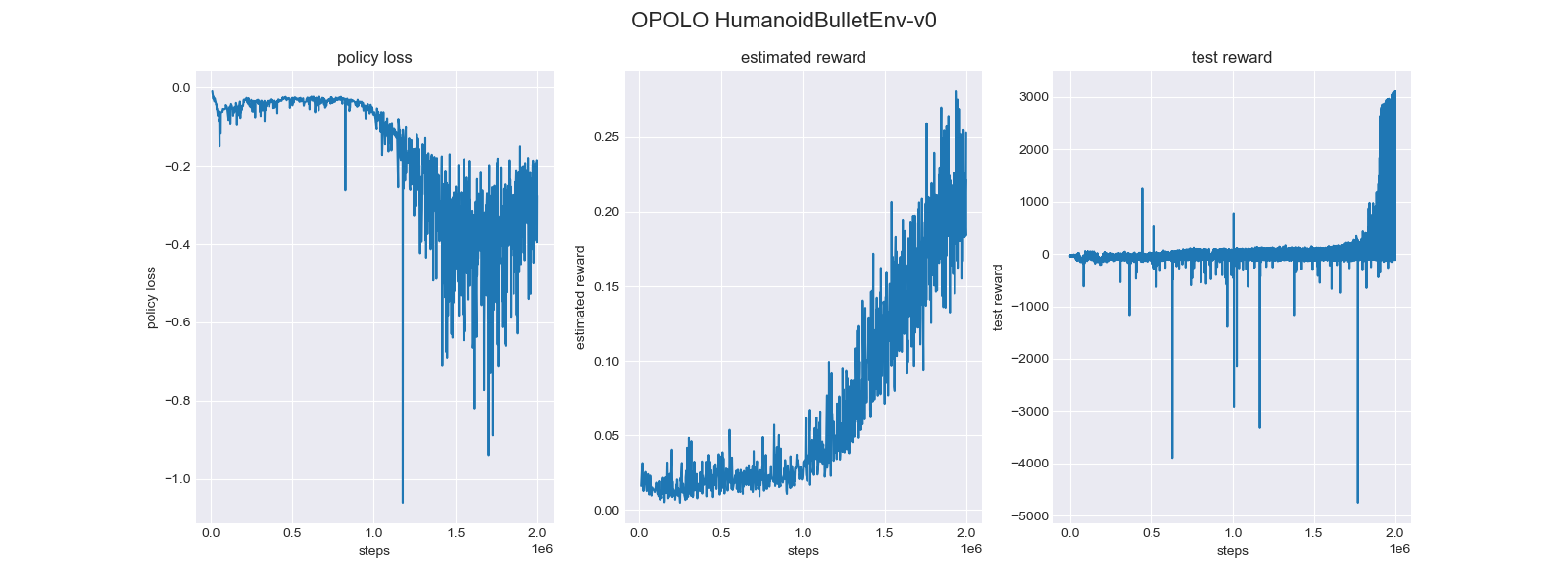}
\includegraphics[width=0.8\textwidth]{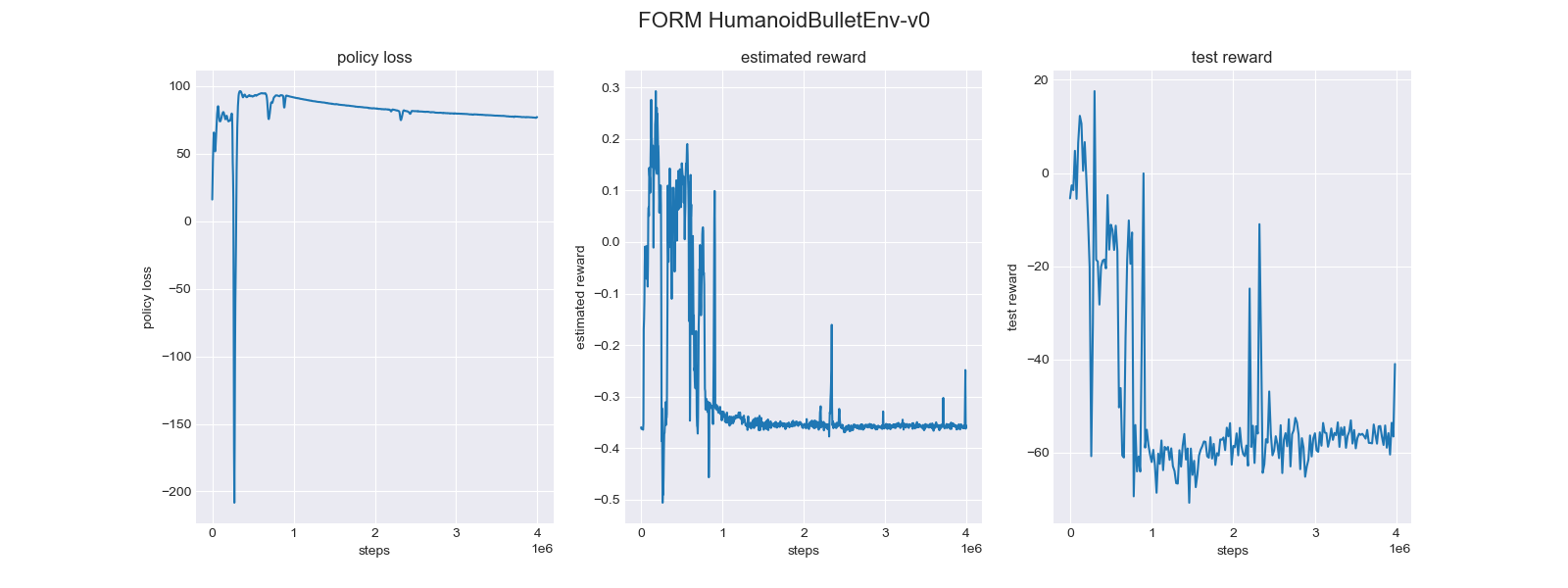}
\includegraphics[width=0.8\textwidth]{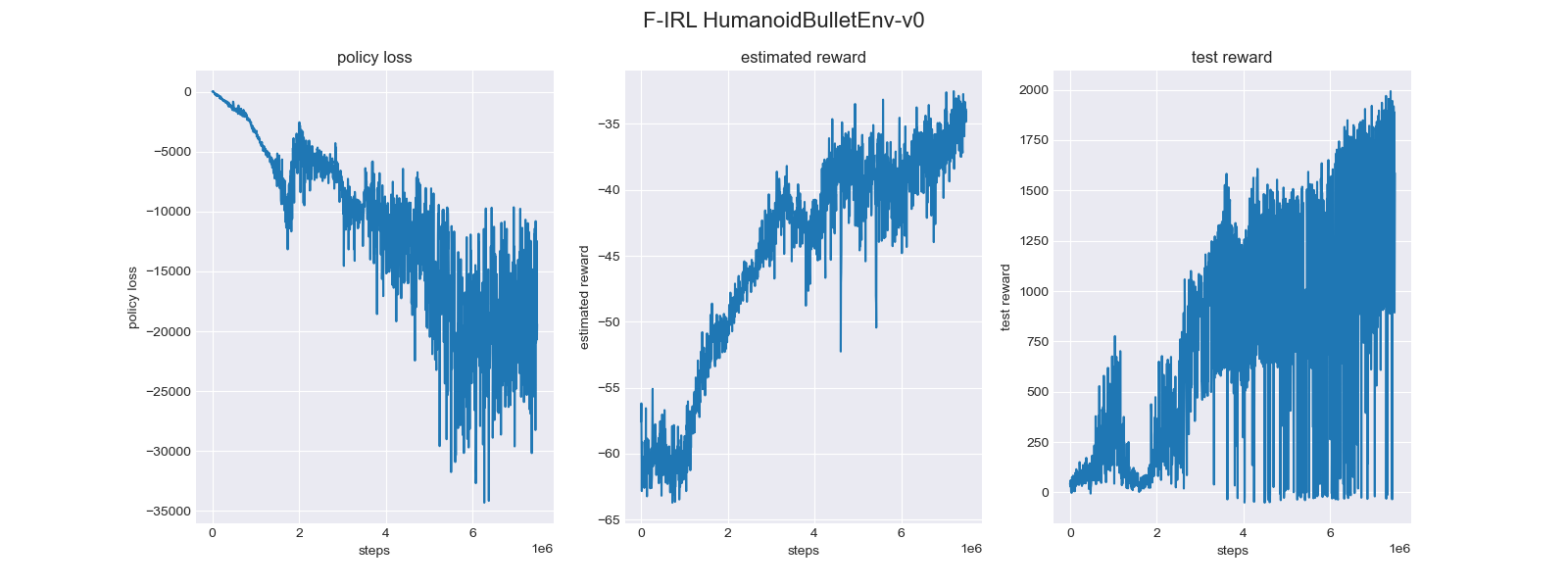}
\caption{The policy loss, estimated reward and the environment test reward during training in the pybullet Humanoid environment using using our proposed SOIL-TDM and the OPOLO, F-IRL, and FORM implementations with 4 expert trajectories.}
\label{fig:tf2}
\end{center}
\end{figure*}

\end{document}